\documentclass[10pt,twocolumn,letterpaper]{article}

\usepackage[pagenumbers]{cvpr} 

\newcommand\blfootnote[1]{%
\begingroup
\renewcommand\thefootnote{}\footnote{#1}%
\addtocounter{footnote}{-1}%
\endgroup
}

\usepackage[utf8]{inputenc}
\usepackage[T1]{fontenc}

\usepackage{color,xcolor}
\usepackage{epsfig}
\usepackage{graphicx}

\usepackage{adjustbox}
\usepackage{array}
\usepackage{booktabs}
\usepackage{colortbl}
\usepackage{float,wrapfig}
\usepackage{hhline}
\usepackage{multirow}
\usepackage{subcaption}
\usepackage[font=small]{caption}
\usepackage{makecell}

\usepackage{amsmath,amsfonts,amsthm,amssymb}
\usepackage{bm}
\usepackage{nicefrac}
\usepackage{microtype}

\usepackage{changepage}
\usepackage{extramarks}
\usepackage{fancyhdr}
\usepackage{lastpage}
\usepackage{setspace}
\usepackage{soul}
\usepackage{xspace}
\usepackage{indentfirst}
\usepackage{pifont}
\usepackage{cuted}
\usepackage{wrapfig}
\definecolor{cvprblue}{rgb}{0.21,0.49,0.74}
\usepackage[pagebackref,breaklinks,colorlinks,allcolors=cvprblue]{hyperref}

\usepackage{url}

\usepackage{algorithm, algorithmic}
\usepackage{enumitem}

\usepackage{wasysym}
\usepackage{duckuments}
\usepackage{pifont}
\usepackage{duckuments}
\usepackage{fancyvrb}
\usepackage{fvextra}

\usepackage{amsthm}  
\theoremstyle{definition} 
\newtheorem{definition}{Definition}[section]  

\theoremstyle{plain} 
\newtheorem{theorem}{Theorem}[section]        

\usepackage{stfloats}

\usepackage{pifont} 

\usepackage{amsmath,amsfonts,bm}

\def\eqref#1{equation~\ref{#1}}

\def\1{\bm{1}}

\DeclareMathAlphabet{\mathsfit}{\encodingdefault}{\sfdefault}{m}{sl}
\SetMathAlphabet{\mathsfit}{bold}{\encodingdefault}{\sfdefault}{bx}{n}

\newcolumntype{L}[1]{>{\raggedright\let\newline\\\arraybackslash\hspace{0pt}}m{#1}}
\newcolumntype{C}[1]{>{\centering\let\newline\\\arraybackslash\hspace{0pt}}m{#1}}
\newcolumntype{R}[1]{>{\raggedleft\let\newline\\\arraybackslash\hspace{0pt}}m{#1}}

\newcommand{\sect}[1]{Section~\ref{sect:#1}}
\newcommand{\app}[1]{Appendix~\ref{app:#1}}

\newcommand{\fig}[1]{Figure~\ref{fig:#1}}

\newcommand{\tbl}[1]{Table~\ref{tab:#1}}

\newcommand{\lblfig}[1]{\label{fig:#1}}
\newcommand{\lblsect}[1]{\label{sect:#1}}
\newcommand{\lblapp}[1]{\label{app:#1}}

\newcommand{\lbltbl}[1]{\label{tab:#1}}

\newcommand{\ignorethis}[1]{}

\renewcommand*{\thefootnote}{\fnsymbol{footnote}}

\makeatletter
\DeclareRobustCommand\onedot{\futurelet\@let@token\@onedot}
\def\@onedot{\ifx\@let@token.\else.\null\fi\xspace}

\makeatother

\definecolor{citecolor}{rgb}{34,139,34}
\definecolor{mydarkblue}{rgb}{0,0.08,1}
\definecolor{mydarkgreen}{rgb}{0.02,0.6,0.02}
\definecolor{mydarkred}{rgb}{0.8,0.02,0.02}
\definecolor{mydarkorange}{rgb}{0.40,0.2,0.02}
\definecolor{mypurple}{RGB}{111,0,255}
\definecolor{myred}{rgb}{1.0,0.0,0.0}
\definecolor{mygold}{rgb}{0.75,0.6,0.12}
\definecolor{mydarkgray}{rgb}{0.66,0.66,0.66}

\newcommand{\myparagraph}[1]{\vspace{0.5pt}\noindent\textbf{#1}\quad}

\def\method{ConvRot\xspace}
\def\engine{ConvLinear4bit\xspace}

\definecolor{mitblue}{rgb}{0.88,0.95,0.96}
\definecolor{mygray}{rgb}{0.9,0.9,0.9}
\definecolor{mypink}{RGB}{175,0,75}
\definecolor{myblue}{RGB}{0,60,160}

\definecolor{cvprblue}{rgb}{0.21,0.49,0.74}
\usepackage[pagebackref,breaklinks,colorlinks,allcolors=cvprblue]{hyperref}

\def\paperID{13040} 
\def\confName{CVPR}
\def\confYear{2026}

\title{ConvRot: Rotation-Based Plug-and-Play 4-bit Quantization for \\Diffusion Transformers}

\author{
{Feice Huang}$^{1\dagger\ast}$,
{Zuliang Han}$^{2\ast}$,
{Xing Zhou}$^{2}$,
{Yihuang Chen}$^{2}$,
{Lifei Zhu}$^{2}$,
{Haoqian Wang}$^{1\textsuperscript{\ding{41}}}$\\
$^1$SIGS, Tsinghua University
$^2$Central Media Technology Institute, Huawei\\
}

\begin{document}
\maketitle

\blfootnote{$^\dagger$Work done during an internship at Central Media Technology Institute.}
\blfootnote{$^\ast$Equal contribution. $\textsuperscript{\ding{41}}$Corresponding author.}

\begin{abstract}
    Diffusion transformers have demonstrated strong capabilities in generating high-quality images. However, as model size increases, the growing memory footprint and inference latency pose significant challenges for practical deployment. Recent studies in large language models (LLMs) show that rotation-based techniques can smooth outliers and enable 4-bit quantization, but these approaches often incur substantial overhead and struggle with row-wise outliers in diffusion transformers. To address these challenges, we propose ConvRot, a group-wise rotation-based quantization method that leverages regular Hadamard transform (RHT) to suppress both row-wise and column-wise outliers while reducing complexity from quadratic to linear. Building on this, we design ConvLinear4bit, a plug-and-play module that integrates rotation, quantization, GEMM, and dequantization, enabling W4A4 inference without retraining and preserving visual quality. Experiments on FLUX.1-dev demonstrate a 2.26$\times$ speedup and 4.05$\times$ memory reduction while maintaining image fidelity. To our knowledge, this is the first application of rotation-based quantization for plug-and-play W4A4 inference in diffusion transformers.

\end{abstract}

\vspace{-5pt}
\section{Introduction}

Diffusion models generate high-fidelity images~\citep{ho2020denoising, rombach2022high}, but scaling their architectures significantly increases memory and inference costs. The recently released Qwen-Image model~\citep{wu2025qwen} reaches a scale of 20B parameters, requiring more than 60 GiB of GPU memory for inference. Quantization is widely used in LLMs to reduce model size and improve inference speed~\citep{zhu2024survey,dettmers2022gpt3,xiao2023smoothquant}, mainly by reducing memory movement and leveraging low-precision compute units in modern GPUs. This makes quantization a promising direction for reducing the memory and latency cost of diffusion models as well. A major source of accuracy loss in quantization comes from outliers, which can distort the scaling factors and degrade performance. Recent studies in LLMs show that rotation-based quantization methods redistribute outliers across channels, enabling 4-bit quantization with minimal accuracy loss~\citep{tseng2024quip,ashkboos2024quarot,liu2024spinquant}, but the extra rotation operations bring non-negligible overhead that offsets part of the speedup. Therefore, the key challenge is to apply rotation-based quantization to diffusion transformers while preserving accuracy and minimizing rotation overhead.

\begin{figure}[b]
  \centering
  \vspace{-10pt}
   \includegraphics[width=1\linewidth]{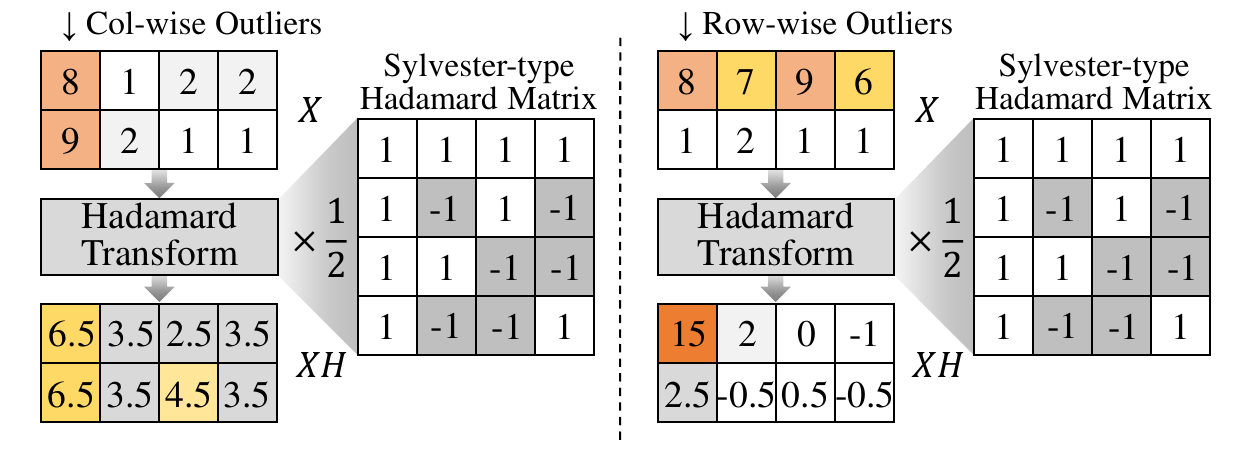}
  \caption{Rotation-based quantization methods using Hadamard matrices can effectively suppress outliers by redistributing energy across channels. However, Sylvester-type Hadamard matrices lead to energy concentration when encountering row-wise outliers.}
   \label{fig:1-1}
\end{figure}

As illustrated in \fig{1-1}, rotation-based quantization methods suppress outliers by applying rotational transformations to both weights and activations. While these methods have been extensively studied in LLMs, directly applying them to diffusion models faces two major challenges. First, the rotation operations themselves introduce substantial computational overhead. Prior works attempt to mitigate this using techniques such as the Fast Walsh-Hadamard Transform (FWHT), which is equivalent to multiplying by Sylvester-type Hadamard matrices, thereby limiting the choice of rotation matrices. They also leverage operation fusion to reduce the need for online rotations~\citep{ashkboos2024quarot}. However, the Adaptive LayerNorm (AdaLN) design in Diffusion Transformers (DiTs)~\citep{peebles2023scalable} can break these fusion strategies, forcing additional online rotations and offsetting the acceleration benefits. Sparse rotation matrices have also been explored~\citep{lin2024duquant}, but without efficient low-level support, they do not lead to practical speedup. Second, as shown in \fig{3-2}, we observe row-wise outliers in certain layers of FLUX.1-dev~\citep{flux1}, which differs from the column-wise outlier patterns commonly found in LLMs. Existing rotation matrix designs, particularly Hadamard matrix-based methods, primarily target column-wise outliers and are therefore ineffective at handling row-wise outliers, leading to noticeable accuracy degradation. Recent work~\citep{federici2025hadanorm} also observed that applying Hadamard directly to DiT can cause degradation due to mean and scale differences across channels.

In this work, we propose ConvRot, a novel rotation-based quantization paradigm. First, we introduce a group-wise rotation scheme that reduces computational complexity from $\mathcal{O}(K^2)$ to $\mathcal{O}(K)$, where $K$ is the number of channels, and allows flexible trade-offs between computation cost and outlier suppression by adjusting the group size $N_0$. 
Second, we adopt the regular Hadamard Transform (RHT) to simultaneously suppress row-wise and column-wise outliers. To support this, we propose a theoretical framework: we formalize the column sum squared property of Hadamard matrices, define the column discrepancy to quantify imbalance, and present a Kronecker-based construction of regular Hadamard matrices for orders that are powers of four, guaranteeing minimal column discrepancy. Based on these regular Hadamard matrices, we implement group-wise RHT with a conv-like matmul operation on weights and activations, which we name ConvRot. Finally, we design ConvLinear4bit, a plug-and-play module that integrates rotation, quantization, GEMM, and dequantization, avoiding expensive loops or extra memory movement while leveraging mature matrix multiplication pipelines on modern GPUs, without requiring complex operator design or additional inference engines. Experimental results demonstrate that ConvRot largely preserves image quality, reduces the memory footprint of the original BF16 DiT by 4.05$\times$, and achieves a 2.26$\times$ speedup on an RTX 4090 24GB.

In summary, our main contributions are:

\begin{itemize}
    \item We provide a theoretical framework for rotation-based quantization: we define the column discrepancy to quantify column sum imbalance, and propose a Kronecker-based construction of regular Hadamard matrices for orders that are powers of four, guaranteeing minimal column discrepancy and therefore particularly effective at mitigating the amplification of row-wise outliers in activations.
    \item We propose ConvRot, a novel group-wise rotation-based quantization paradigm that leverages regular Hadamard Transform (RHT) to simultaneously smooth row-wise and column-wise outliers, reducing computational complexity from $\mathcal{O}(N^2)$ to $\mathcal{O}(N)$ and significantly lowering latency compared to global rotations.
    \item We design ConvLinear4bit, a plug-and-play module that integrates rotation, quantization, GEMM, and dequantization, enabling training-free W4A4 inference for all linear layers in diffusion models. 
    \item To the best of our knowledge, we are the first to achieve training-free, fully 4-bit (W4A4) inference on diffusion transformers with acceptable accuracy degradation, without auxiliary high-precision branches and while avoiding major visual artifacts.
\end{itemize}
\section{Related Work}

\begin{figure}[b]
    \centering
    \includegraphics[width=\linewidth]{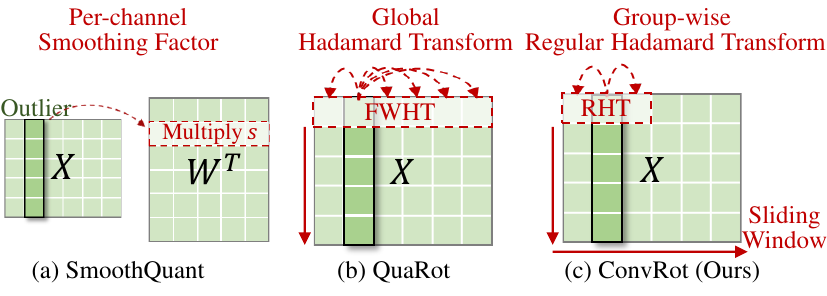}
  \caption{Illustration of how different transformations redistribute outliers. 
  (a) \textbf{SmoothQuant}~\citep{xiao2023smoothquant}: per-channel diagonal transform $T(\mathbf{X}) = \mathbf{X}\,\mathrm{diag}(\mathbf{s})^{-1}$ shifts activation outlier magnitudes into the corresponding channel weights. 
  (b) \textbf{QuaRot}~\citep{ashkboos2024quarot}: global Hadamard transform $T(\mathbf{X}) = \mathbf{X}\mathbf{H}$ with orthogonal $\mathbf{H}$ evenly redistributes activation energy. 
  (c) \textbf{ConvRot (Ours)}: Group-wise Regular Hadamard Transform performs local smoothing of activations within sliding windows.}
    \lblfig{3-1}
  \end{figure}
  
\subsection{Quantization for LLMs}

Quantization reduces memory traffic and computation by adopting low-precision formats, while also enabling efficient use of hardware-specific accelerators such as INT4 tensor cores~\citep{frantar-gptq, lin2024awq, dettmers2022gpt3}. However, naive per-tensor or per-channel post-training quantization (PTQ) schemes suffer from outliers that dominate the dynamic range, leading to substantial accuracy degradation. Rotation-based quantization addresses this by applying orthogonal transforms to distribute outliers across channels, producing smoother distributions with fewer extreme values~\citep{tseng2024quip}. Yet, these rotations introduce quadratic complexity, which offsets the potential acceleration. Prior works mitigate the cost with fast hadamard transforms~\citep{ashkboos2024quarot}, fuse the rotation into adjacent linear layers~\citep{liu2024spinquant}, or block-diagonal rotations~\citep{lin2024duquant}. While effective for LLMs, these designs face challenges in diffusion models, fusion breaks under adaptive normalization layers ~\citep{peebles2023scalable}, and block-diagonal rotations fail to deliver speedup proportional to their reduced computation. In contrast, our ConvRot employs a lightweight group-wise rotation that reduces complexity to linear while preserving sufficient smoothing, and can be directly applied to diffusion models without architectural changes.

\subsection{Acceleration of Diffusion Models}

Diffusion models~\citep{ho2020denoising} achieve state-of-the-art performance in image and video generation~\citep{wu2025qwen, kong2024hunyuanvideo}, but their inference speed remains a major limitation for deployment due to the inherently slow and computationally intensive iterative process. Existing acceleration strategies include few-step samplers~\citep{song2020denoising,lu2022dpm, lu2022dpm++}, distillation~\citep{salimans2021progressive, luo2023latent, yin2024one}, pruning~\citep{zhao2024dynamic}, and caching~\citep{liu2025timestep}. Recently, quantization has also been explored for diffusion models~\citep{li2023q, zhao2024mixdq, zhao2024vidit, li2024svdquant}. However, unlike language models, where latency is often dominated by weight loading, diffusion models are computationally bounded~\citep{li2024svdquant}. As a result, weight-only quantization is insufficient for diffusion models, both weights and activations must be quantized to fully exploit low-precision hardware. However, existing methods either maintain activations in higher precision~\citep{dettmers2023qlora}, preventing the use of low-precision tensor cores, or rely on customized inference engines~\citep{li2024svdquant}, which complicates deployment. 
By contrast, our method supports end-to-end 4-bit weight-activation quantization, fully exploiting low-precision hardware units. Furthermore, the proposed ConvLinear4bit layer is plug-and-play, requiring no specialized inference engine, and integrates seamlessly with existing quantized operators to deliver both memory reduction and practical speedup.

\section{Preliminary}

\begin{figure*}[t]
    \centering
    \includegraphics[width=\linewidth]{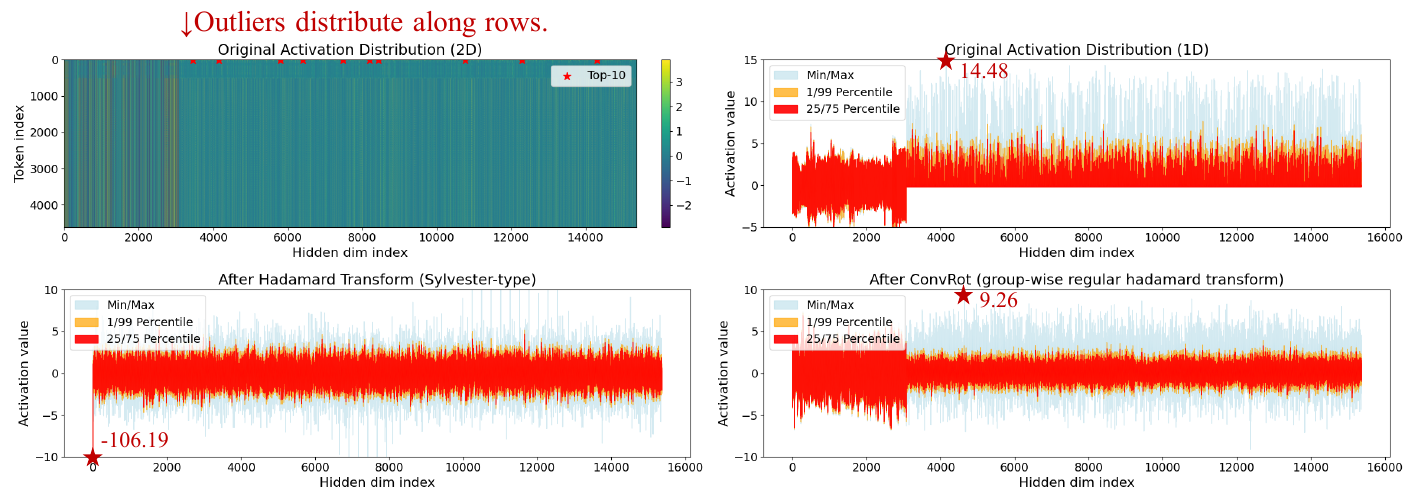}
	\caption{Effect of Hadamard transforms on the \texttt{single\_transformer\_blocks.37.proj\_out} activations in Flux. The standard transform amplifies outliers (max $=106.19$), while the group-wise regular transform suppresses them (max $=9.26$), compared to the original (max $=14.48$).}
    \lblfig{3-2}
     \vspace{-5pt}
\end{figure*}

\subsection{Equivalent Transformation Based Quantization Methods}
\lblsect{3-1}
Given an activation vector $\mathbf{x} \in \mathbb{R}^n$, a uniform $b$-bit quantizer is defined as
\begin{equation}
    Q(\mathbf{x}) = \mathrm{round}\!\left( \frac{\mathbf{x}}{s} \right),
\end{equation}
where $s$ is a scaling factor. Large-magnitude outliers in $\mathbf{x}$ inflate $s$, reducing the effective resolution for most elements and making low-bit quantization challenging.

Transformation-based quantization methods apply an orthogonal or diagonal transformation $T(\cdot)$ to redistribute activation magnitudes before quantization while preserving the computation. For a linear layer $\mathbf{Y} = \mathbf{X} \mathbf{W}^\top$, this invariance is expressed as
\begin{equation}
    \mathbf{X} \mathbf{W}^\top = T(\mathbf{X}) \, T'(\mathbf{W})^\top,
\end{equation}
where $T'(\cdot)$ is the corresponding transform on the weight. As illustrated in \fig{3-1}, different choices of $T(\cdot)$ correspond to different ways of redistributing outliers. The core challenge is to design $T(\cdot)$ to \textbf{suppress outlier amplitudes with minimal computational overhead}, enabling smaller scaling factors $s$ and higher effective precision in low-bit quantization.

\subsection{Regular Hadamard Matrix for Rotation}

\begin{definition}[Hadamard Matrix ($\mathcal{H}$-Matrix)]
A Hadamard matrix (abbrev. $\mathcal{H}$-Matrix) $\mathbf{H}_n \in \{\pm 1\}^{n \times n}$ satisfies
\begin{equation}
    \mathbf{H}_n \mathbf{H}_n^\top = n \mathbf{I}_n,
\end{equation}
where $\mathbf{I}_n$ is the $n \times n$ identity. Normalized by $1/\sqrt{n}$, $\mathbf{H}_n$ is orthogonal, making it well-suited for rotation-based quantization since orthogonality redistributes outliers. Empirically, $\mathcal{H}$-Matrix rotations outperform random orthogonal ones~\citep{liu2024spinquant, tseng2024quip}.
\end{definition}

A common construction of Hadamard matrices is the Sylvester-type recursion, which produces a standard Hadamard matrix with the first row and column filled with 1s:

\[
\mathbf{H}_1 = [1], \quad
\mathbf{H}_{2n} =
\begin{bmatrix}
\mathbf{H}_n & \mathbf{H}_n \\
\mathbf{H}_n & -\mathbf{H}_n
\end{bmatrix}.
\]

The Fast Walsh-Hadamard Transform (FWHT) exploits this structure to reduce the matrix-vector multiplication complexity from $\mathcal{O}(n^2)$ to $\mathcal{O}(n \log n)$, using only additions and subtractions. Recent work~\citep{lin2024duquant} demonstrates that block-wise rotations can preserve most of the benefits while reducing computational cost. It is worth noting that the FWHT is equivalent to multiplying by a Sylvester-type Hadamard matrix, \emph{whose first column is all ones}, which can inadvertently amplify row-wise outliers in the activations, as illustrated in \fig{3-2}.

\begin{theorem}[Column Sum Squared Property]
For an $\mathcal{H}$-Matrix $\mathbf{H}_n$,
\begin{equation}
    \sum_{j=1}^n \left( \sum_{i=1}^n \mathbf{H}_{ij} \right)^2 = n^2.
\end{equation}
\end{theorem}

\begin{definition}[Column Discrepancy]
For an $\mathcal{H}$-Matrix $\mathbf{H}_n$, the \emph{column discrepancy} is defined as
\begin{equation}
    \|\mathbf{H}^\top \mathbf{1}\|_\infty = \max_j \Big|\sum_i H_{ij}\Big|,
\end{equation}
where $\mathbf{1}$ is the all-ones vector. It measures the largest deviation of a column sum from zero.
\end{definition}

This metric is related to \emph{combinatorial discrepancy}~\citep{spencer1985six, matousek1999geometric}, defined for $A \in \{\pm 1\}^{n \times m}$ as
\begin{equation}
    \mathrm{disc}(A) = \min_{\varepsilon \in \{\pm 1\}^n} \|A^\top \varepsilon\|_\infty.
\end{equation}
Here $\varepsilon$ is a $\pm 1$ coloring of rows, chosen to minimize the maximum column imbalance.  
In our case, the column discrepancy corresponds to the fixed coloring $\varepsilon = \mathbf{1}$, giving a natural upper bound on $\mathrm{disc}(\mathbf{H})$.

For $\mathcal{H}$-Matrices, the column discrepancy always satisfies
\begin{equation}
    \sqrt{n} \;\le\; \|\mathbf{H}^\top \mathbf{1}\|_\infty \;\le\; n.
\end{equation}
The lower bound follows from the column sum squared property, while the upper bound is achieved by Sylvester-type matrices that contain identical columns. This motivates the study of \emph{regular} $\mathcal{H}$-Matrices, which attain the minimum value.

\begin{definition}[Regular $\mathcal{H}$-Matrix]
An $\mathcal{H}$-Matrix is \emph{regular} if each row and column sums to $\pm \sqrt{n}$.
\end{definition}

\begin{theorem} \label{thm:regular-discrepancy}
Regular $\mathcal{H}$-Matrices attain the minimal possible column discrepancy:
\begin{equation}
    \max_{j} \Big| \sum_{i=1}^n \mathbf{H}_{ij} \Big| = \sqrt{n}.
\end{equation}
\end{theorem}

\begin{theorem}[Kronecker Construction]  \label{thm:kronecker}
For every $k \ge 1$, a regular $\mathcal{H}$-Matrix of order $n = 4^k$ exists. Starting from
\begin{equation}
    \mathbf{H}_4 = 
    \begin{bmatrix}
        1 & 1 & 1 & -1 \\
        1 & 1 & -1 & 1 \\
        1 & -1 & 1 & 1 \\
        -1 & 1 & 1 & 1
    \end{bmatrix},
\end{equation}
one obtains $\mathbf{H}_{4^{k+1}} = \mathbf{H}_{4^k} \otimes \mathbf{H}_4$ via the Kronecker product. Each $\mathbf{H}_{4^k}$ remains regular. 
\end{theorem}

All proofs are deferred to \app{proof}.
  
We leverage this property to design a \emph{group-wise regular $\mathcal{H}$-Matrix rotation} scheme. It reduces peak activations and latency while preserving the smoothing benefits of $\mathcal{H}$-Matrix rotations, making it practical for large diffusion models where global rotations are expensive and may amplify outliers.

\begin{figure*}[t]
    \centering
    \includegraphics[width=\linewidth]{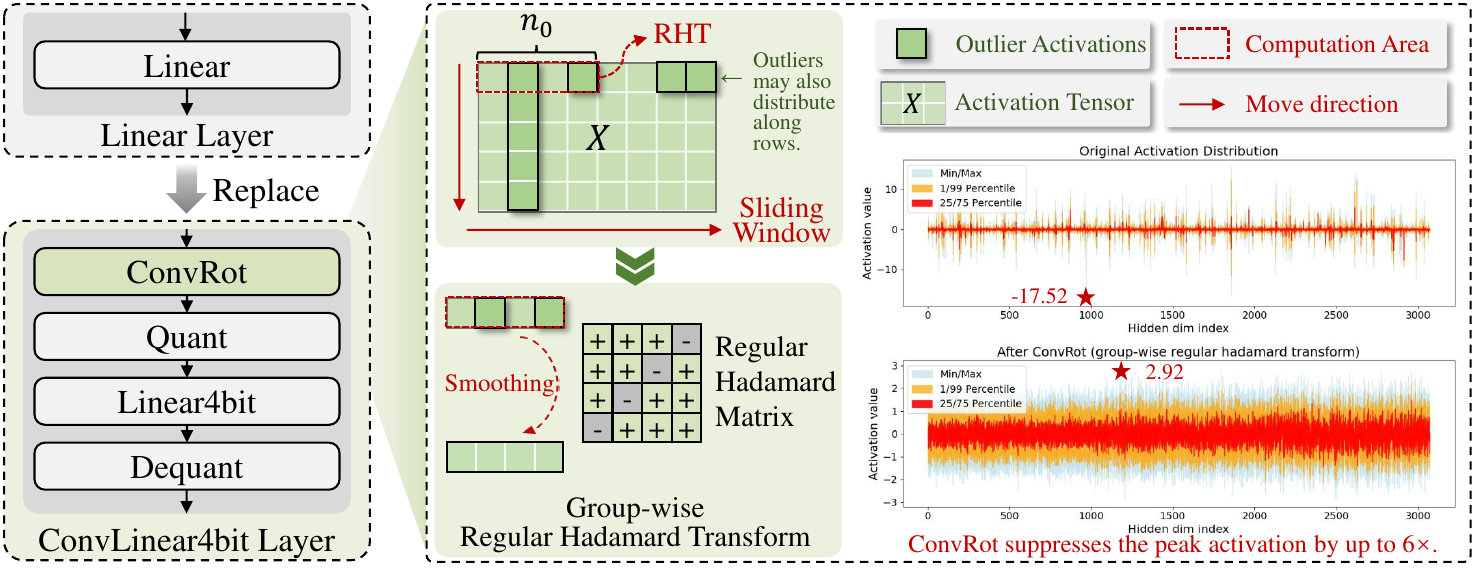}
	\caption{
		Overview of \method. Left: \engine serves as a plug-and-play replacement for Linear layers. Right: \method applies Regular Hadamard Transform (RHT) on non-overlapping sliding windows of the activation tensor, with each window multiplied by a regular Hadamard matrix.
    }
    \lblfig{4-1}
\end{figure*}
\section{Method}
In this work, we present an plug-and-play W4A4 quantization method for Diffusion Transformers (DiTs). As illustrated in \fig{4-1}, Our approach consists of two core components: \textbf{\method} and \textbf{\engine}. \method performs group-wise regular Hadamard rotations, where the group-wise design reduces computational cost and the regular structure prevents row-wise outlier aggregation, smoothing activation and weight distributions before quantization. \engine integrates \method, 4-bit quantization, matrix multiplication, and dequantization into a single linear layer, enabling straightforward plug-and-play replacement of the original layers. By simply replacing the original linear layers with \engine, we can perform low-precision inference on large-scale DiT models while maintaining high visual quality.

\subsection{Motivation}
Our first motivation comes from the high computational cost of existing rotation-based quantization methods, such as QuaRot~\citep{ashkboos2024quarot} and SpinQuant~\citep{liu2024spinquant}, which typically apply a single global Hadamard rotation of order $n$. Because this approach redistributes outliers across all channels, it incurs quadratic complexity $\mathcal{O}(K^2)$, making it expensive for large-scale diffusion models. To reduce this cost, we propose limiting the rotation to smaller groups of size $N_0$, allowing a flexible trade-off between computational overhead and the effectiveness of outlier smoothing.

The second motivation stems from a property of large Hadamard matrices with high-magnitude columns. As observed in FLUX (see \fig{3-2}), when outliers in activations are distributed along specific rows, multiplying by such a matrix can inadvertently concentrate these outliers, increasing their magnitude instead of smoothing them. This row-wise aggregation arises mainly from the combination of large matrix size and column-sums with large absolute values. Applying Hadamard rotations on smaller, group-wise blocks alleviates this issue by limiting the rotation scope, while using regular Hadamard matrices further prevents the emergence of large-magnitude column-sums, effectively smoothing activations and avoiding outlier concentration.

These two insights motivate the design of \method: (i) reducing computational cost by restricting the rotation scope, and (ii) preventing row-wise outlier concentration through the use of regular Hadamard matrices.

\subsection{\method: Group-Wise Regular Hadamard Rotation}

As discussed in \sect{3-1}, inserting a Hadamard transformation within a matrix multiplication can be interpreted as rotating the input or weight space without changing the output distribution. Building upon this, we propose \method, which applies a \textbf{group-wise Regular Hadamard Transform (RHT)} to control the scope of rotation and improve both computational efficiency and outlier handling.

Existing rotation-based quantization methods, such as QuaRot and SpinQuant, typically apply a global Hadamard transform of order $K$, which incurs quadratic complexity $\mathcal{O}(K^2)$ since outliers are redistributed across all channels simultaneously. This becomes prohibitive for large-scale diffusion models. Our key insight is that by partitioning the feature dimension into blocks of size $N_0$ and applying a \textbf{Regular Hadamard Transform} within each block, we can reduce the computational cost and localize outlier redistribution.

Formally, given a standard linear layer
\begin{equation}
    \mathbf{Y} = \mathbf{X} \mathbf{W}^\top, \quad \mathbf{X}\in \mathbb{R}^{M\times K}, \mathbf{W}\in \mathbb{R}^{N\times K},
\end{equation}
we partition the input and weight matrices into column-wise blocks of size $N_0$:

\begin{align}
    \mathbf{X} &= [\mathbf{X}_1, \mathbf{X}_2, \dots, \mathbf{X}_{\lceil K/N_0 \rceil}], \\
    \mathbf{W} &= [\mathbf{W}_1, \mathbf{W}_2, \dots, \mathbf{W}_{\lceil K/N_0 \rceil}].
\end{align}

For each block, we insert a Regular Hadamard Rotation (RHT):
\begin{equation}
    \mathbf{Y} = \sum_{i=1}^{\lceil K/N_0 \rceil} \text{RHT}(\mathbf{X}_i) \, \text{RHT}(\mathbf{W}_i)^\top.
\end{equation}

By performing group-wise RHT, we reduce computational complexity from $\mathcal{O}(K^2)$ to $\mathcal{O}(K)$, while preserving effective outlier suppression. Importantly, the equivalence property ensures that this local rotation does not change the overall linear transformation; it only redistributes the information within each block, providing finer-grained control over activation distributions. \tbl{5-1} shows the Outlier Amplitude, defined as $\max |XH|$ (the maximum absolute value among all elements of $XH$), under different rotation types and group sizes $N_0$.

Existing Hadamard-based quantization methods typically rely on the FWHT implementation from Dao AI Lab~\citep{hadamard_transform_package}. This implementation follows FFT-like butterfly operations and, in its current form, primarily runs on CUDA cores without using Tensor Cores. In our work, we implement \method using matrix multiplication, which reduces additional memory movement and benefits from the optimized matmul pipelines on modern GPUs. This leads to noticeable speedups compared with FWHT-based approaches. Formally, a group rotation of size $N_0$ can be expressed as a convolution-like operation on the input activation with kernel size $[1, N_0]$, channels equal to $N_0$, and stride $(1, N_0)$, which motivates the name \textbf{\method} as it integrates the ideas of convolution and rotation.

\subsection{\engine}

Building on \method, we develop \textbf{\engine}, which allows straightforward replacement of original linear layers for plug-and-play 4-bit inference as shown in \fig{4-1}. \engine integrates \method, quantization, 4-bit matrix multiplication, and dequantization into a single layer. 

For \method, we implement the group-wise Hadamard rotation via reshape-based matrix multiplication, minimizing additional memory movement. The quantization, 4-bit matrix multiplication, and dequantization operations follow the design in QuaRot~\citep{ashkboos2024quarot} and utilize highly optimized CUDA kernels to take advantage of GPUs' int4 Tensor Cores.

In summary, \method provides a flexible mechanism to control the number of channels participating in outlier redistribution, trading off computation and smoothing effectiveness. Matmul-based implementation allows leveraging modern GPU pipelines efficiently, while \engine enables practical, plug-and-play 4-bit weight-activation quantization for large-scale diffusion models, achieving significant memory reduction and inference speedup without sacrificing image quality. 
\section{Experiments}

\begin{table}[b]
    \centering
    \caption{Outlier amplitude after Hadamard rotation across different transformer layers. 
The two layers respectively represent cases with and without significant row-wise outlier patterns. 
Additional results are provided in the \app{layer}}
    \footnotesize
    \setlength{\tabcolsep}{1pt}
    
    \begin{tabular}{l|cccccc}
    \toprule
    \multicolumn{7}{c}{\textit{single\_transformer\_blocks.0.proj\_out}} \\
    \midrule
    \multirow{2}{*}{\makecell[l]{Hadamard\\Type}} & \multicolumn{6}{c}{Outlier Amplitude $\downarrow$} \\
    \cmidrule(lr){2-7}
    & $N_0\!=\!16$ & $N_0\!=\!64$ & $N_0\!=\!256$ & $N_0\!=\!1024$ & Global & Original \\
    \midrule
    Random   & 6.74{\color{mypink}\scriptsize -46\%} & 6.47{\color{mypink}\scriptsize -48\%} & 5.23{\color{mypink}\scriptsize -58\%} & \underline{3.96}{\color{mypink}\scriptsize -68\%} & & 12.43 \\
    Standard & 6.05{\color{mypink}\scriptsize -51\%} & 6.48{\color{mypink}\scriptsize -48\%} & 8.66{\color{myblue}\scriptsize +20\%}  & 16.45{\color{myblue}\scriptsize +32\%} & 49.59{\color{myblue}\scriptsize +299\%} & 12.43 \\
    \rowcolor{gray!12}
    Regular  & 6.54{\color{mypink}\scriptsize -47\%} & 5.57{\color{mypink}\scriptsize -55\%} & 4.75{\color{mypink}\scriptsize -62\%} & \textbf{3.68}{\color{mypink}\scriptsize -70\%} & & 12.43 \\
    \midrule
    \multicolumn{7}{c}{\textit{single\_transformer\_blocks.0.attn\_to\_k}} \\
    \midrule
    \multirow{2}{*}{\makecell[l]{Hadamard\\Type}} & \multicolumn{6}{c}{Outlier Amplitude $\downarrow$} \\
    \cmidrule(lr){2-7}
    & $N_0\!=\!16$ & $N_0\!=\!64$ & $N_0\!=\!256$ & $N_0\!=\!1024$ & Global & Original \\
    \midrule
    Random   & 18.09{\color{mypink}\scriptsize -57\%} & 12.70{\color{mypink}\scriptsize -69\%} & 8.59{\color{mypink}\scriptsize -79\%} & 5.66{\color{mypink}\scriptsize -86\%} & & 41.62 \\
    Standard & 11.16{\color{mypink}\scriptsize -73\%} & 6.32{\color{mypink}\scriptsize -85\%} & 4.29{\color{mypink}\scriptsize -90\%} & \underline{4.14}{\color{mypink}\scriptsize -90\%} & 4.21{\color{mypink}\scriptsize -89\%} & 41.62 \\
    \rowcolor{gray!12}
    Regular  & 10.95{\color{mypink}\scriptsize -74\%} & 6.46{\color{mypink}\scriptsize -84\%} & 4.48{\color{mypink}\scriptsize -89\%} & \textbf{3.54}{\color{mypink}\scriptsize -92\%} & & 41.62 \\
    \bottomrule
    \end{tabular}
    
    \lbltbl{5-1}
\end{table}

\begin{table*}[t]
    \renewcommand{\arraystretch}{1}
    \footnotesize
    \setlength{\tabcolsep}{8pt}
    \centering

\caption{
    End-to-end performance comparison across transformer models.
    Lower LPIPS/FID and higher PSNR/IR indicate better performance.
    SVDQuant maintains a parallel 16-bit LoRA branch, while our mixed-precision strategy randomly executes a subset of layers in INT8.
}

\lbltbl{5-2}
\begin{tabular}{ccccccccccc}
    \toprule
    & & & \multicolumn{4}{c}{MJHQ} & \multicolumn{4}{c}{sDCI} \\
    \cmidrule(lr){4-7} \cmidrule(lr){8-11}
    Model & Precision & Method & \multicolumn{2}{c}{Quality} & \multicolumn{2}{c}{Similarity} & \multicolumn{2}{c}{Quality} & \multicolumn{2}{c}{Similarity} \\
    \cmidrule(lr){4-5} \cmidrule(lr){6-7} \cmidrule(lr){8-9} \cmidrule(lr){10-11}
    & & & FID$\downarrow$ & IR$\uparrow$ & LPIPS$\downarrow$ & PSNR$\uparrow$ & FID$\downarrow$ & IR$\uparrow$ & LPIPS$\downarrow$ & PSNR$\uparrow$ \\

    \midrule
    \multirow{6}{*}{\makecell{FLUX.1\\-dev\\(50 Steps)}} 
    & BF16 & -- & 10.07 & 0.99 & -- & -- & 13.83 & 1.05 & -- & -- \\

    \cmidrule{2-11}
    & \cellcolor{mygray}INT W8A8 & \cellcolor{mygray}Ours & \cellcolor{mygray}9.81 & \cellcolor{mygray}0.98& \cellcolor{mygray}0.13 & \cellcolor{mygray}22.62 & \cellcolor{mygray}13.46 & \cellcolor{mygray}1.05 & \cellcolor{mygray}0.15 & \cellcolor{mygray}22.59 \\

    \cmidrule{2-11}
    & W4A16 & NF4 & 12.10 & 0.96 & 0.23 & 19.23 &14.09 & 0.99 & 0.24 & 17.89 \\
    & INT W4A4 {\scriptsize +16bit LoRA} & SVDQuant & 10.01 & 0.97 & 0.18 & 20.49 & 13.71 & 1.03 & 0.20 & 19.65 \\
    \cmidrule{2-11}

    & \cellcolor{mygray}INT W4A4 & \cellcolor{mygray}Ours & \cellcolor{mygray}12.32 & \cellcolor{mygray}0.84 & \cellcolor{mygray}0.22 & \cellcolor{mygray}19.43 & \cellcolor{mygray}16.01 & \cellcolor{mygray}0.87 & \cellcolor{mygray}0.25 & \cellcolor{mygray}17.66 \\
    & \cellcolor{mygray}INT W4A4{\scriptsize +20\% 8bit Mixed} & \cellcolor{mygray}Ours & \cellcolor{mygray}10.03 & \cellcolor{mygray}0.97 & \cellcolor{mygray}0.18& \cellcolor{mygray}20.73 & \cellcolor{mygray}14.00 & \cellcolor{mygray}1.01 & \cellcolor{mygray}0.21 & \cellcolor{mygray}19.41 \\

    \midrule
    \multirow{6}{*}{\makecell{FLUX.1\\-schnell\\(4 Steps)}} 
    & BF16 & -- & 11.59 & 0.91 & -- & -- & 11.01 & 0.97 & -- & -- \\

    \cmidrule{2-11}
    & \cellcolor{mygray}INT W8A8 & \cellcolor{mygray}Ours & \cellcolor{mygray}10.86 & \cellcolor{mygray}0.96 & \cellcolor{mygray}0.15 & \cellcolor{mygray}21.35 & \cellcolor{mygray}11.09 & \cellcolor{mygray}0.99 & \cellcolor{mygray}0.16 & \cellcolor{mygray}19.91 \\

    \cmidrule{2-11}
    & W4A16 & NF4 & 11.52 & 0.91 & 0.20 & 18.03 & 10.93 & 0.96 & 0.28 & 16.77 \\
    & INT W4A4 {\scriptsize +16bit LoRA} & SVDQuant & 11.47 & 0.92 & 0.21 & 18.07 & 10.47 & 1.02 & 0.26 & 19.66 \\
    \cmidrule{2-11}
    & \cellcolor{mygray}INT W4A4 & \cellcolor{mygray}Ours & \cellcolor{mygray}13.38 & \cellcolor{mygray}0.81 & \cellcolor{mygray}0.23 & \cellcolor{mygray}17.5 & \cellcolor{mygray}12.37 & \cellcolor{mygray}0.84 & \cellcolor{mygray}0.30 & \cellcolor{mygray}16.98 \\
    & \cellcolor{mygray}INT W4A4 {\scriptsize +20\% 8bit Mixed} & \cellcolor{mygray}Ours & \cellcolor{mygray}11.48 & \cellcolor{mygray}0.92 & \cellcolor{mygray}0.20 & \cellcolor{mygray}18.11 & \cellcolor{mygray}11.13 & \cellcolor{mygray}0.97 & \cellcolor{mygray}0.23 & \cellcolor{mygray}18.71 \\

    \bottomrule
\end{tabular}
\end{table*}

\subsection{Setups}

\myparagraph{Models.}
We conduct our experiments on FLUX.1-dev and FLUX.1-schnell~\citep{flux1}, which are both 12B-parameter text-to-image diffusion models known for their high-quality image generation capabilities. Typically, inference with these models requires over 30GiB of GPU memory, making it hard to deploy on consumer-grade hardware.

\myparagraph{Datasets.}
Following \citet{li2024svdquant}, we evaluate generation quality on a subset of 5K prompts stratified sampled across categories from the MJHQ-30K dataset~\citep{li2024playground} and the
summarized Densely Captioned Images (sDCI)~\citep{urbanek2024picture} datasets.

\begin{figure*}[t]
    \captionsetup{font=small}
    \centering
    \includegraphics[width=\textwidth]{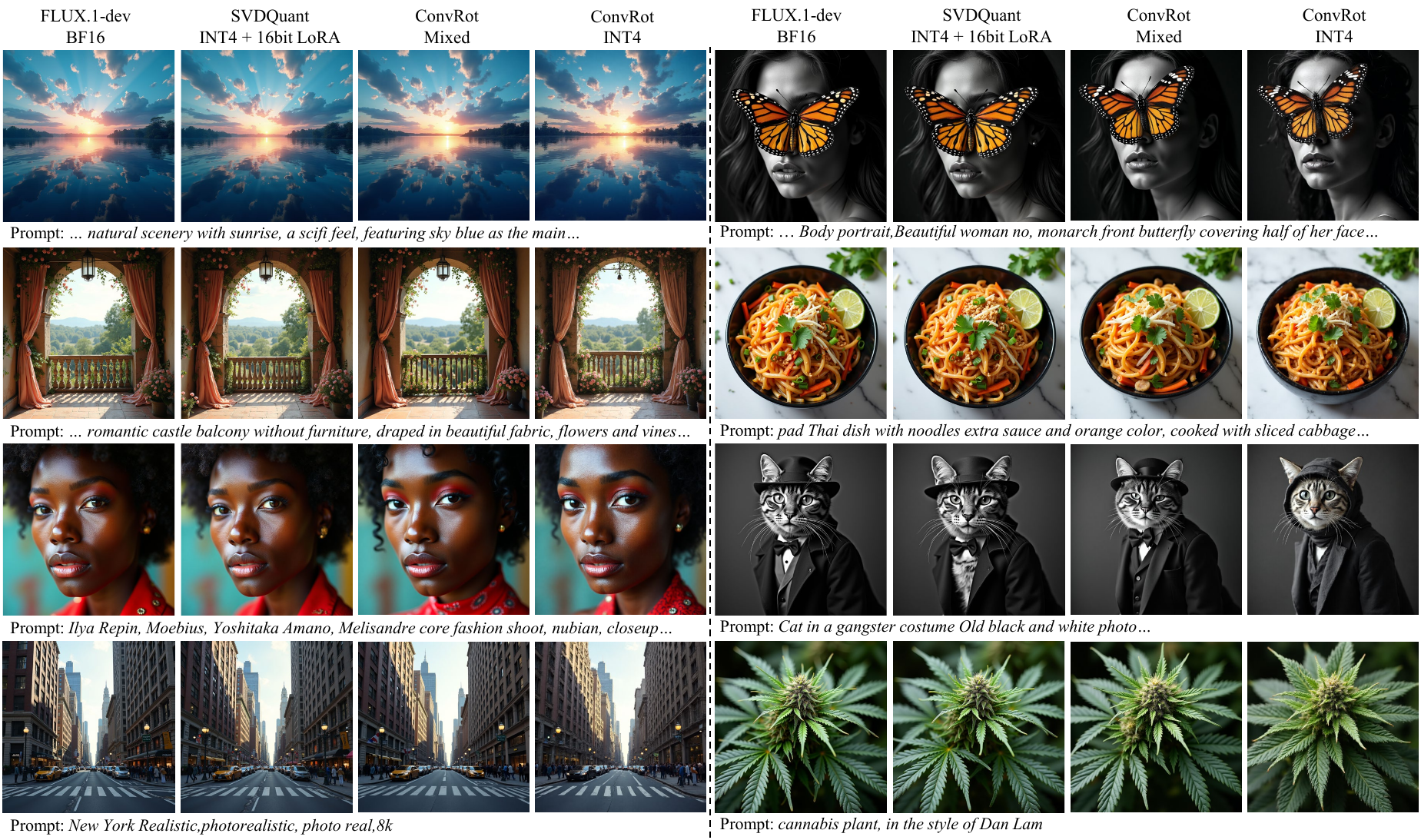}
    \caption{Visual comparison of our method using different rotation sizes on the MJHQ-30K dataset. Prompts cover diverse themes including food, human portraits, animals, landscapes, indoor scenes, and figurines.}

    \lblfig{visual}
    \end{figure*}

\myparagraph{Baselines.}
We compare \method against three representative baselines: BF16, SVDQuant~\citep{li2024svdquant}, QuaRot~\citep{ashkboos2024quarot}, and NF4~\citep{dettmers2023qlora}. SVDQuant leverages a hybrid 16-bit branch to achieve high-quality 4-bit inference; QuaRot applies rotation-based quantization originally designed for LLMs; and NF4 provides information-theoretically optimal 4-bit quantization primarily for training or LoRA finetuning rather than efficient inference.
We provide a detailed comparison of these methods in \app{compare}.

\myparagraph{Metrics.} We conduct a comprehensive evaluation at both the single-layer and end-to-end levels.
For single linear layer analysis, we evaluate \textbf{Precision} by measuring the post-rotation outlier amplitude and the maximum layer error. We also assess \textbf{Efficiency} by benchmarking the rotation latency and overall layer latency.
For the end-to-end text-to-image task, we evaluate from two aspects: \textbf{Quality}, using FID~\citep{heusel2017gans} (lower is better) and ImageReward (IR)~\citep{xu2024imagereward} (higher is better); and \textbf{Similarity} to the original BF16 model, using LPIPS (lower is better) and PSNR (higher is better). We also report DiT memory footprint and end-to-end generation latency.

\subsection{Single-Layer Analysis}

\tbl{5-1} compares precision and efficiency of different rotation implementations in some single Flux linear layers with the prompt ``A cute cat.''. Outlier Amplitude denotes the maximum absolute activation, Rotation Latency the rotation runtime, and Layer Latency the full layer runtime. As shown in \fig{3-2}, layers such as \texttt{single\_transformer\_blocks.\{i\}.proj\_out} and \texttt{transformer\_blocks.\{i\}.ff\_context.net.2} exhibit pronounced row-wise outliers, with QuaRot's official Sylvester-type Hadamard (size 15360) reaching Outlier Amplitude 105.63.

\textbf{Implementation details.} 
For \method, we use per-token/per-channel 4-bit quantization with regular Hadamard rotations, denoting \method-$N_0$ as group size $N_0$. Layers with larger outlier amplitudes use larger rotations ($N_0=1024$) to better smooth outliers. SVDQuant and QuaRot use their official defaults. Experiments run on a single RTX 4090 (24GB), with CPU offloading for models exceeding GPU memory.

For the FWHT implementation, small group sizes (e.g., $N_0=16$) require many rotation calls, increasing latency, while larger $N_0$ aggregates row-wise outliers due to the first column of all ones, reducing precision. In contrast, Group-wise RHT shows decreasing Outlier Amplitude with larger $N_0$, effectively mitigating row-wise outliers. Rotation latency also decreases with $N_0$, reaching 1.246ms at $N_0=64$ (1.55$\times$ speedup over FP16). Considering both precision and efficiency, we select $N_0=256$ as the default settings for ConvRot. For a more detailed performance analysis of FWHT and RHT, please refer to the \app{speed}.

\subsection{End-to-End Performance on Text-to-Image Generation}
\lblsect{5-3}

\begin{table}[t]
    \renewcommand{\arraystretch}{0.8}
    \caption{DiT Memory and Latency comparison on FLUX.1-dev (50 steps) on a single 4090 GPU.}
    \lbltbl{5-3-mem-lat}
    \scriptsize 
    \setlength{\tabcolsep}{2mm} 
    \centering
    \begin{tabular}{l l | c c}
        \toprule
        Method & Precision & DiT Memory (GiB) & Latency (s) \\
        \midrule
        Baseline & BF16 & 22.7 & 54.6 \\
        NF4 & W4A16 & 6.9 & 38.6 \\
        SVDQuant & W4A4 + 16 bit LoRA & 6.5 & 14.9 \\
        \midrule
        \method & W4A4 & 5.6 & 23.2 \\
        & W4A4 + INT8 Mixed & 7.0 & 28.3 \\
        \bottomrule
    \end{tabular}
    \vspace{-5pt}
\end{table}

We compare our ConvRot approach with SVDQuant and the BF16 baseline on FLUX.1-dev and FLUX.1-Schnell. As summarized in \tbl{5-2}, we evaluate both \textbf{Similarity} (LPIPS, PSNR) and \textbf{Quality} (FID, IR). Lower LPIPS/FID scores and higher PSNR/IR scores indicate better performance.

\method significantly reduces both memory footprint and latency compared to the BF16 baseline, while introducing only modest degradation in image quality. Metrics such as FID, IR, LPIPS, and PSNR indicate that the quality drop mainly arises from reduced model representation capacity under full INT4 inference, rather than from row-wise outliers. Nevertheless, \method still achieves state-of-the-art visual quality among current INT4 methods, producing images with perceptually acceptable fidelity.

\begin{figure}[b]
    \captionsetup{font=small}
    \centering
    \includegraphics[width=0.5\textwidth]{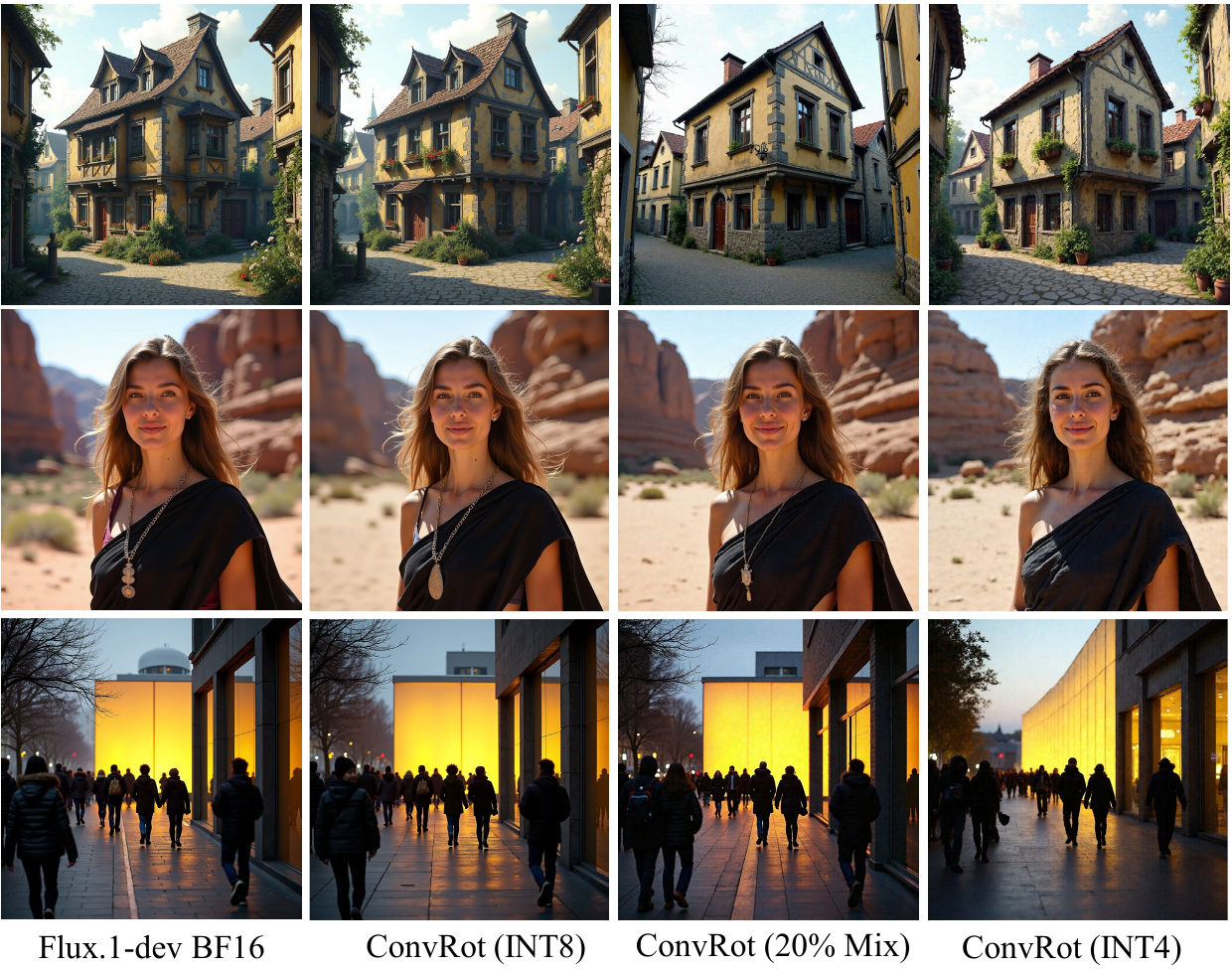}
    \caption{Qualitative impact of hybrid-precision inference on sDCI.}

    \lblfig{5-2}
    \end{figure}

To improve image quality, we adopt a hybrid-precision strategy in which 20\% of the layers are empirically selected for INT8 execution, while the rest remain in INT4. Details are provided in \app{mix}. This configuration restores fine-grained details and perceptual quality, closely approaching the performance of SVDQuant with its 16-bit LoRA branch. As shown in \tbl{5-2}, \method achieves a strong balance between memory efficiency, latency, and image fidelity. \fig{5-2} illustrates that introducing a fraction of INT8 layers enhances texture sharpness and global coherence compared with full INT4 inference.

\subsection{Ablation Study}

We further investigate the impact of group size $N_0$ and rotation type on image generation quality. Unlike the end-to-end experiments in \sect{5-3}, here we apply the same $N_0$ uniformly across all layers. This setup isolates the effect of rotation configuration on the model's representational capacity and image fidelity.

\begin{table}[t]
    \renewcommand{\arraystretch}{0.8}
    \caption{Ablation on the choice of precision, rotation type, and size. More results can be found in \tbl{app-3}.}
    \lbltbl{5-4}
    \centering
    \footnotesize
    \setlength{\tabcolsep}{3pt}
    \begin{tabular}{cccccccc}
    \toprule
     &  & \multicolumn{2}{c}{Rotation} & \multicolumn{4}{c}{MJHQ} \\
    \cmidrule(lr){3-4} \cmidrule(lr){5-8}
    \multirow{-2}{*}{Precision} & \multirow{-2}{*}{Mixed} & Type & Size & FID$\downarrow$ & IR$\uparrow$ & LPIPS$\downarrow$ & PSNR$\uparrow$ \\
    \midrule
    BF16 &  & -- & -- & 10.07 & 0.993 & -- & -- \\
    \midrule
    INT W8A8 &  & Standard & 256 & 9.93 & 0.981 & 0.140 & 22.64 \\
    INT W8A8 &  & Standard & 1024 & 9.91 & 0.985 & 0.139 & 22.61 \\
    \rowcolor{mygray}INT W8A8 &  & Regular & 256 & 9.82 & 0.981 & 0.131 & 22.62 \\
    \rowcolor{mygray}INT W8A8 &  & Regular & 1024 & 9.77 & 0.977 & 0.129 & 22.66 \\
    \midrule
    {\color[HTML]{BFBFBF} INT W4A4} & {\color[HTML]{BFBFBF}\ding{51}} & {\color[HTML]{BFBFBF} Standard} & {\color[HTML]{BFBFBF} Full} & {\color[HTML]{BFBFBF} 98.04} & {\color[HTML]{BFBFBF} -2.240} & {\color[HTML]{BFBFBF} 0.766} & {\color[HTML]{BFBFBF} 7.70} \\
    INT W4A4 &  & Standard & 256 & 12.55 & 0.835 & 0.234 & 19.23 \\
    INT W4A4 &  & Standard & 1024 & 13.86 & 0.808 & 0.286 & 18.85 \\
    INT W4A4 & \ding{51} & Standard & 256 & 12.45 & 0.861 & 0.193 & 20.17 \\
    INT W4A4 & \ding{51} & Standard & 1024 & 13.29 & 0.850 & 0.290 & 18.45 \\
    \rowcolor{mygray}INT W4A4 &  & Regular & 256 & 12.32 & 0.841 & 0.220 & 19.43 \\
    \rowcolor{mygray}INT W4A4 &  & Regular & 1024 & 12.30 & 0.855 & 0.215 & 19.52 \\
    \rowcolor{mygray}INT W4A4 & \ding{51} & Regular & 256 & 10.03 & 0.973 & 0.186 & 20.73 \\
    \rowcolor{mygray}INT W4A4 & \ding{51} & Regular & 1024 & 10.02 & 0.983 & 0.184 & 20.71 \\
    \bottomrule
    \end{tabular}
\end{table}

More ablation results are provided in the \tbl{app-3}.

\section{Conclusion}
We present ConvRot, a novel rotation-based quantization framework for diffusion transformers that enables efficient W4A4 INT4 inference while preserving image quality. To address the challenges of row-wise and column-wise outliers in activations, we introduce a group-wise rotation scheme based on regular Hadamard matrices, reducing computational complexity from quadratic to linear and significantly lowering rotation latency compared to global rotations. Building on this, we design ConvLinear4bit, a plug-and-play module that fuses rotation, quantization, GEMM, and dequantization, allowing all linear layers in a diffusion model to be quantized without retraining. Extensive experiments on FLUX.1-dev demonstrate that our approach stably suppresses outliers, reduces memory usage by 4.05$\times$, and achieves a 2.26$\times$ speedup, while maintaining high-fidelity image generation. To our knowledge, this work is the first to apply rotation-based quantization to diffusion transformers for fully INT4 W4A4 inference, providing a practical solution for accelerating large-scale text-to-image generation with minimal quality loss.
{
    \small
    \bibliographystyle{ieeenat_fullname}
    \bibliography{main}

@String(CVPR= {IEEE Conf. Comput. Vis. Pattern Recog.})

@String(ICCV= {Int. Conf. Comput. Vis.})

@String(ECCV= {Eur. Conf. Comput. Vis.})

@String(ICLR = {Int. Conf. Learn. Represent.})

@String(CVPR  = {CVPR})

@String(ICCV  = {ICCV})

@String(ECCV  = {ECCV})

@String(ICLR  = {ICLR})

@STRING{CVPR = {CVPR}}

@STRING{ICCV = {ICCV}}

@STRING{ECCV = {ECCV}}

@STRING{ICLR = {ICLR}}

@STRING{ICML = {ICML}}

@STRING{NeurIPS = {NeurIPS}}

@STRING{MLSys = {MLSys}}

@inproceedings{song2020denoising,
  title={Denoising Diffusion Implicit Models},
  author={Song, Jiaming and Meng, Chenlin and Ermon, Stefano},
  booktitle=ICLR,
  year={2020}
}

@inproceedings{salimans2021progressive,
  title={Progressive Distillation for Fast Sampling of Diffusion Models},
  author={Salimans, Tim and Ho, Jonathan},
  booktitle=ICLR,
  year={2021}
}

@article{ho2020denoising,
  title={Denoising diffusion probabilistic models},
  author={Ho, Jonathan and Jain, Ajay and Abbeel, Pieter},
  journal=NeurIPS,
  year={2020}
}

@article{heusel2017gans,
  title={Gans trained by a two time-scale update rule converge to a local nash equilibrium},
  author={Heusel, Martin and Ramsauer, Hubert and Unterthiner, Thomas and Nessler, Bernhard and Hochreiter, Sepp},
  journal=NeurIPS,
  year={2017}
}

@inproceedings{rombach2022high,
  title={High-resolution image synthesis with latent diffusion models},
  author={Rombach, Robin and Blattmann, Andreas and Lorenz, Dominik and Esser, Patrick and Ommer, Bj{\"o}rn},
  booktitle=CVPR,
  year={2022}
}

@inproceedings{lu2022dpm,
  title={Dpm-solver: A fast ode solver for diffusion probabilistic model sampling in around 10 steps},
  author={Lu, Cheng and Zhou, Yuhao and Bao, Fan and Chen, Jianfei and Li, Chongxuan and Zhu, Jun},
  booktitle=NeurIPS,
  year={2022}
}

@article{lu2022dpm++,
  title={Dpm-solver++: Fast solver for guided sampling of diffusion probabilistic models},
  author={Lu, Cheng and Zhou, Yuhao and Bao, Fan and Chen, Jianfei and Li, Chongxuan and Zhu, Jun},
  journal={arXiv preprint arXiv:2211.01095},
  year={2022}
}

@inproceedings{li2023q,
  title={Q-diffusion: Quantizing diffusion models},
  author={Li, Xiuyu and Liu, Yijiang and Lian, Long and Yang, Huanrui and Dong, Zhen and Kang, Daniel and Zhang, Shanghang and Keutzer, Kurt},
  booktitle=ICCV,
  year={2023}
}

@article{luo2023latent,
  title   = {Latent Consistency Models: Synthesizing High-Resolution Images with Few-Step Inference},
  author  = {Simian Luo and Yiqin Tan and Longbo Huang and Jian Li and Hang Zhao},
  year    = {2023},
  journal = {arXiv preprint arXiv: 2310.04378}
}

@inproceedings{xiao2023smoothquant,
  title={Smoothquant: Accurate and efficient post-training quantization for large language models},
  author={Xiao, Guangxuan and Lin, Ji and Seznec, Mickael and Wu, Hao and Demouth, Julien and Han, Song},
  booktitle=ICML,
  year={2023},
}

@inproceedings{lin2024awq,
  title={AWQ: Activation-aware Weight Quantization for On-Device LLM Compression and Acceleration},
  author={Lin, Ji and Tang, Jiaming and Tang, Haotian and Yang, Shang and Chen, Wei-Ming and Wang, Wei-Chen and Xiao, Guangxuan and Dang, Xingyu and Gan, Chuang and Han, Song},
  booktitle=MLSys,
  year={2024}
}

@article{dettmers2022gpt3,
  title={Gpt3. int8 (): 8-bit matrix multiplication for transformers at scale},
  author={Dettmers, Tim and Lewis, Mike and Belkada, Younes and Zettlemoyer, Luke},
  journal=NeurIPS,
  year={2022}
}

@inproceedings{zhao2024mixdq,
  title={Mixdq: Memory-efficient few-step text-to-image diffusion models with metric-decoupled mixed precision quantization},
  author={Zhao, Tianchen and Ning, Xuefei and Fang, Tongcheng and Liu, Enshu and Huang, Guyue and Lin, Zinan and Yan, Shengen and Dai, Guohao and Wang, Yu},
  booktitle=ECCV,
  year={2024},
}

@article{zhao2024vidit,
  title={ViDiT-Q: Efficient and Accurate Quantization of Diffusion Transformers for Image and Video Generation},
  author={Zhao, Tianchen and Fang, Tongcheng and Liu, Enshu and Rui, Wan and Soedarmadji, Widyadewi and Li, Shiyao and Lin, Zinan and Dai, Guohao and Yan, Shengen and Yang, Huazhong and others},
  journal={arXiv preprint arXiv:2406.02540},
  year={2024}
}

@article{frantar-gptq,
  title={{GPTQ}: Accurate Post-training Compression for Generative Pretrained Transformers}, 
  author={Elias Frantar and Saleh Ashkboos and Torsten Hoefler and Dan Alistarh},
  year={2023},
  journal=ICLR
}

@inproceedings{peebles2023scalable,
  title={Scalable diffusion models with transformers},
  author={Peebles, William and Xie, Saining},
  booktitle=ICCV,
  year={2023}
}

@inproceedings{yin2024one,
  title={One-step diffusion with distribution matching distillation},
  author={Yin, Tianwei and Gharbi, Micha{\"e}l and Zhang, Richard and Shechtman, Eli and Durand, Fredo and Freeman, William T and Park, Taesung},
  booktitle=CVPR,
  year={2024}
}

@misc{li2024playground,
      title={Playground v2.5: Three Insights towards Enhancing Aesthetic Quality in Text-to-Image Generation}, 
      author={Daiqing Li and Aleks Kamko and Ehsan Akhgari and Ali Sabet and Linmiao Xu and Suhail Doshi},
      year={2024},
      eprint={2402.17245},
      archivePrefix={arXiv},
      primaryClass={cs.CV}
}

@inproceedings{urbanek2024picture,
  title={A picture is worth more than 77 text tokens: Evaluating clip-style models on dense captions},
  author={Urbanek, Jack and Bordes, Florian and Astolfi, Pietro and Williamson, Mary and Sharma, Vasu and Romero-Soriano, Adriana},
  booktitle=CVPR,
  year={2024}
}

@article{ashkboos2024quarot,
  title={Quarot: Outlier-free 4-bit inference in rotated llms},
  author={Ashkboos, Saleh and Mohtashami, Amirkeivan and Croci, Maximilian and Li, Bo and Cameron, Pashmina and Jaggi, Martin and Alistarh, Dan and Hoefler, Torsten and Hensman, James},
  journal=NeurIPS,
  year={2024}
}

@article{liu2024spinquant,
  title={SpinQuant--LLM quantization with learned rotations},
  author={Liu, Zechun and Zhao, Changsheng and Fedorov, Igor and Soran, Bilge and Choudhary, Dhruv and Krishnamoorthi, Raghuraman and Chandra, Vikas and Tian, Yuandong and Blankevoort, Tijmen},
  journal={arXiv preprint arXiv:2405.16406},
  year={2024}
}

@inproceedings{
    dettmers2023qlora,
    title={{QL}o{RA}: Efficient Finetuning of Quantized {LLM}s},
    author={Tim Dettmers and Artidoro Pagnoni and Ari Holtzman and Luke Zettlemoyer},
    booktitle=NeurIPS,
    year={2023},
}

@misc{flux1,
  author = "Black-Forest-Labs",
  title = "FLUX.1",
  year = "2024",
  url = "https://blackforestlabs.ai/",
}

@article{xu2024imagereward,
  title={Imagereward: Learning and evaluating human preferences for text-to-image generation},
  author={Xu, Jiazheng and Liu, Xiao and Wu, Yuchen and Tong, Yuxuan and Li, Qinkai and Ding, Ming and Tang, Jie and Dong, Yuxiao},
  journal=NeurIPS,
  year={2024}
}

@article{wu2025qwen,
  title={Qwen-image technical report},
  author={Wu, Chenfei and Li, Jiahao and Zhou, Jingren and Lin, Junyang and Gao, Kaiyuan and Yan, Kun and Yin, Sheng-ming and Bai, Shuai and Xu, Xiao and Chen, Yilei and others},
  journal={arXiv preprint arXiv:2508.02324},
  year={2025}
}

@article{tseng2024quip,
  title={Quip\#: Even better llm quantization with hadamard incoherence and lattice codebooks},
  author={Tseng, Albert and Chee, Jerry and Sun, Qingyao and Kuleshov, Volodymyr and De Sa, Christopher},
  journal={arXiv preprint arXiv:2402.04396},
  year={2024}
}

@article{zhu2024survey,
  title={A survey on model compression for large language models},
  author={Zhu, Xunyu and Li, Jian and Liu, Yong and Ma, Can and Wang, Weiping},
  journal={Transactions of the Association for Computational Linguistics},
  volume={12},
  pages={1556--1577},
  year={2024},
  publisher={MIT Press 255 Main Street, 9th Floor, Cambridge, Massachusetts 02142, USA~…}
}

@article{li2024svdquant,
  title={Svdquant: Absorbing outliers by low-rank components for 4-bit diffusion models},
  author={Li, Muyang and Lin, Yujun and Zhang, Zhekai and Cai, Tianle and Li, Xiuyu and Guo, Junxian and Xie, Enze and Meng, Chenlin and Zhu, Jun-Yan and Han, Song},
  journal={arXiv preprint arXiv:2411.05007},
  year={2024}
}

@article{lin2024duquant,
  title={Duquant: Distributing outliers via dual transformation makes stronger quantized llms},
  author={Lin, Haokun and Xu, Haobo and Wu, Yichen and Cui, Jingzhi and Zhang, Yingtao and Mou, Linzhan and Song, Linqi and Sun, Zhenan and Wei, Ying},
  journal={Advances in Neural Information Processing Systems},
  volume={37},
  pages={87766--87800},
  year={2024}
}

@article{kong2024hunyuanvideo,
  title={Hunyuanvideo: A systematic framework for large video generative models},
  author={Kong, Weijie and Tian, Qi and Zhang, Zijian and Min, Rox and Dai, Zuozhuo and Zhou, Jin and Xiong, Jiangfeng and Li, Xin and Wu, Bo and Zhang, Jianwei and others},
  journal={arXiv preprint arXiv:2412.03603},
  year={2024}
}

@article{zhao2024dynamic,
  title={Dynamic diffusion transformer},
  author={Zhao, Wangbo and Han, Yizeng and Tang, Jiasheng and Wang, Kai and Song, Yibing and Huang, Gao and Wang, Fan and You, Yang},
  journal={arXiv preprint arXiv:2410.03456},
  year={2024}
}

@inproceedings{liu2025timestep,
  title={Timestep Embedding Tells: It's Time to Cache for Video Diffusion Model},
  author={Liu, Feng and Zhang, Shiwei and Wang, Xiaofeng and Wei, Yujie and Qiu, Haonan and Zhao, Yuzhong and Zhang, Yingya and Ye, Qixiang and Wan, Fang},
  booktitle={Proceedings of the Computer Vision and Pattern Recognition Conference},
  pages={7353--7363},
  year={2025}
}

@article{spencer1985six,
  title={Six standard deviations suffice},
  author={Spencer, Joel},
  journal={Transactions of the American mathematical society},
  volume={289},
  number={2},
  pages={679--706},
  year={1985}
}

@book{matousek1999geometric,
  title={Geometric discrepancy: An illustrated guide},
  author={Matousek, Jiri},
  volume={18},
  year={1999},
  publisher={Springer Science \& Business Media}
}

@article{federici2025hadanorm,
  title={HadaNorm: Diffusion Transformer Quantization through Mean-Centered Transformations},
  author={Federici, Marco and Del Chiaro, Riccardo and van Breugel, Boris and Whatmough, Paul and Nagel, Markus},
  journal={arXiv preprint arXiv:2506.09932},
  year={2025}
}

@misc{hadamard_transform_package,
  title        = {hadamard–transform: Fast Walsh–Hadamard Transform (FWHT) implementation in PyTorch},
  author       = {Portnoy, Amit},
  howpublished = {\url{https://pypi.org/project/hadamard-transform/}},
  note         = {Version\,0.1.3, released July 5, 2022},
  year         = {2022}
}

@article{agarwal2024hadacore,
  title={Hadacore: Tensor core accelerated hadamard transform kernel},
  author={Agarwal, Krish and Astra, Rishi and Hoque, Adnan and Srivatsa, Mudhakar and Ganti, Raghu and Wright, Less and Chen, Sijia},
  journal={arXiv preprint arXiv:2412.08832},
  year={2024}
}
}

\clearpage
\setcounter{page}{1}
\maketitlesupplementary

\section{Proofs}
\lblapp{proof}
\subsection{Proof of Theorem 3.1 (Column Sum Squared Property)}
\lblapp{appendix:proof-column-sum}
\begin{proof}
Let $\mathbf{H}_n$ be a Hadamard matrix of order $n$, satisfying $\mathbf{H}_n \mathbf{H}_n^\top = n \mathbf{I}_n$.  
Define the column sums as $c_j = \sum_{i=1}^n H_{ij}$ for $j=1,\dots,n$.  
We consider the squared $\ell_2$ norm of the vector of column sums:
\begin{equation}
    \sum_{j=1}^n c_j^2 = \sum_{j=1}^n \left(\sum_{i=1}^n H_{ij}\right)^2.
\end{equation}
This can be expressed as
\begin{equation}
    \sum_{j=1}^n c_j^2 = \|\mathbf{H}_n^\top \mathbf{1}\|_2^2 = \mathbf{1}^\top \mathbf{H}_n \mathbf{H}_n^\top \mathbf{1}.
\end{equation}
Using the orthogonality property $\mathbf{H}_n \mathbf{H}_n^\top = n \mathbf{I}_n$, we obtain
\begin{equation}
    \mathbf{1}^\top \mathbf{H}_n \mathbf{H}_n^\top \mathbf{1} = n \cdot \mathbf{1}^\top \mathbf{1} = n \cdot n = n^2.
\end{equation}
Therefore,
\begin{equation}
    \sum_{j=1}^n \left( \sum_{i=1}^n H_{ij} \right)^2 = n^2,
\end{equation}
which proves the claim.
\end{proof}

\subsection{Proof of Theorem 3.2 (Column Discrepancy of Regular Hadamard)}
\lblapp{proof-regular-min}
\begin{proof}
By definition, the column discrepancy of $\mathbf{H}_n$ is
\begin{equation}
    \|\mathbf{H}_n^\top \mathbf{1}\|_\infty = \max_{1 \le j \le n} \left|\sum_{i=1}^n H_{ij}\right|.
\end{equation}
From \app{appendix:proof-column-sum}, we know that the squared column sums satisfy
\begin{equation}
    \sum_{j=1}^n \left(\sum_{i=1}^n H_{ij}\right)^2 = n^2.
\end{equation}
If $\mathbf{H}_n$ is regular, each column sum satisfies $\sum_i H_{ij} = \pm \sqrt{n}$, so the maximum absolute column sum is exactly $\sqrt{n}$.  
Since $\sqrt{n}$ is also the theoretical minimum discrepancy achievable by any Hadamard matrix, regular Hadamard matrices attain the optimum.
\end{proof}

\subsection{Proof of Theorem 3.3 (Kronecker Construction of Regular Hadamard Matrices)}

\lblapp{proof-kronecker}
\begin{proof}
We prove by induction on $k$.

\textbf{Base case:} For $k=1$, the given $4\times4$ matrix $\mathbf{H}_4$ is regular, since each row and column sums to $\pm 2 = \pm \sqrt{4}$.

\textbf{Inductive step:} Assume $\mathbf{H}_{4^k}$ is a regular Hadamard matrix, i.e., each row and column sums to $\pm \sqrt{4^k}$.  
Consider $\mathbf{H}_{4^{k+1}} = \mathbf{H}_{4^k} \otimes \mathbf{H}_4$.  
For any column of $\mathbf{H}_{4^{k+1}}$, the Kronecker product structure ensures that its entries are composed of four blocks, each proportional to a column of $\mathbf{H}_{4^k}$.  
Therefore, the column sum is
\begin{equation}
    \sum_{i} H^{(4^{k+1})}_{ij} = \left(\sum_{u} H^{(4^k)}_{u,v}\right) \cdot \left(\sum_{w} H^{(4)}_{w,z}\right),
\end{equation}
where $v,z$ index the corresponding columns in $\mathbf{H}_{4^k}$ and $\mathbf{H}_4$.  
By the induction hypothesis, $\sum_{u} H^{(4^k)}_{u,v} = \pm \sqrt{4^k}$, and since $\mathbf{H}_4$ is regular, $\sum_{w} H^{(4)}_{w,z} = \pm 2$.  
Thus,
\begin{equation}
    \sum_{i} H^{(4^{k+1})}_{ij} = (\pm \sqrt{4^k})(\pm 2) = \pm \sqrt{4^{k+1}}.
\end{equation}
Hence $\mathbf{H}_{4^{k+1}}$ is also regular. By induction, a regular Hadamard matrix exists for all $n=4^k$.
\end{proof}

\section{Efficiency Analysis}
\lblapp{speed}

In this section, we compare the efficiency of the FWHT implementation from Dao AI Lab~\citep{hadamard_transform_package} with our proposed group-wise RHT.\footnote{For some reason, the experiments in this section were conducted on an A100 GPU, rather than the RTX 4090 used in the main paper.} It is worth noting that the butterfly operations in FWHT correspond to the Sylvester-type Hadamard transform, which restricts FWHT from being applied to modified Hadamard matrices.

From a theoretical perspective, applying a global Hadamard transform directly via matrix multiplication incurs a quadratic cost of $\mathcal{O}(K^2)$ when the number of channels is $K$. QuaRot~\citep{ashkboos2024quarot} decomposes $K$ as $K = 2^n m$, where the $2^n$ part can be accelerated using FWHT with complexity $\mathcal{O}(K n)$, giving an overall complexity of $\mathcal{O}(K(m+n))$.
From an engineering standpoint, however, the widely used FWHT implementation from Dao AI Lab relies on FFT-like butterfly operators and does not utilize Tensor Cores. While butterfly operations can in principle be reformulated into small matrix multiplications to make use of Tensor Core acceleration~\citep{agarwal2024hadacore}, this optimization has not yet seen broad adoption in existing methods, so we do not include its performance in our comparisons.

Although ConvLinear4bit integrates rotation, quantization, INT4 GEMM, and dequantization into a unified module, we do not fuse these operations at the kernel level. Instead, all stages directly reuse the existing implementations from QuaRot~\citep{ashkboos2024quarot}. As shown in \fig{app-3}, we benchmark both FWHT and RHT under group sizes of 64, 256, and 1024 using a linear layer with
$W \in \mathbb{R}^{3072 \times 15360}$,
$X \in \mathbb{R}^{1 \times 4608 \times 15360}$,
and $Y \in \mathbb{R}^{1 \times 4608 \times 3072}$.
The only performance difference between the two methods comes from the rotation stage. Because FWHT cannot parallelize the normalization step ($1/\sqrt{n}$), an additional Hadamard normalization must be applied, further increasing its latency. For the global FWHT case, since the size 15360 does not appear in QuaRot's precomputed Hadamard list, the transform must be implemented by combining FWHT with matrix multiplication; both this matmul and the subsequent normalization are executed using FP16 GEMM kernels, and we report their combined latency.

As shown in \tbl{app-1}, FWHT exhibits poor hardware utilization when the group size is small. Similarly, group-wise RHT also suffers from low occupancy at a group size of 64, but achieves strong performance at the more commonly used group sizes of 256 and 1024. Although group-wise RHT becomes slightly slower than FWHT at a group size of 1024 due to its larger computational load, this gap can be reduced with further kernel optimization, which is beyond the scope of this paper and left for future work.

\begin{table}[t]
\centering
\caption{Profiling statistics of FWHT kernels with different group sizes (Rotation Latency).}
\lbltbl{app-1}
\scriptsize
\setlength{\tabcolsep}{1mm} 
\begin{tabular}{lcccccccc}
\toprule
Group Size & Shared Mem & Blocks/SM & Warps/SM & Occupancy & Wall Dur. (ms) \\
\midrule
64  & 256  & 10240 & 2560   & 13\%  & 0.822 \\
256  & 1024 & 2560  & 2560  & 50\%  & 0.209 \\
1024 & 4096 & 640   & 2560 & 100\% & 0.229 \\
\bottomrule
\end{tabular}
\end{table}

\begin{figure}[t]
  \centering
   \vspace{-5pt}
  \hspace{-20pt} 
  \includegraphics[width=\linewidth]{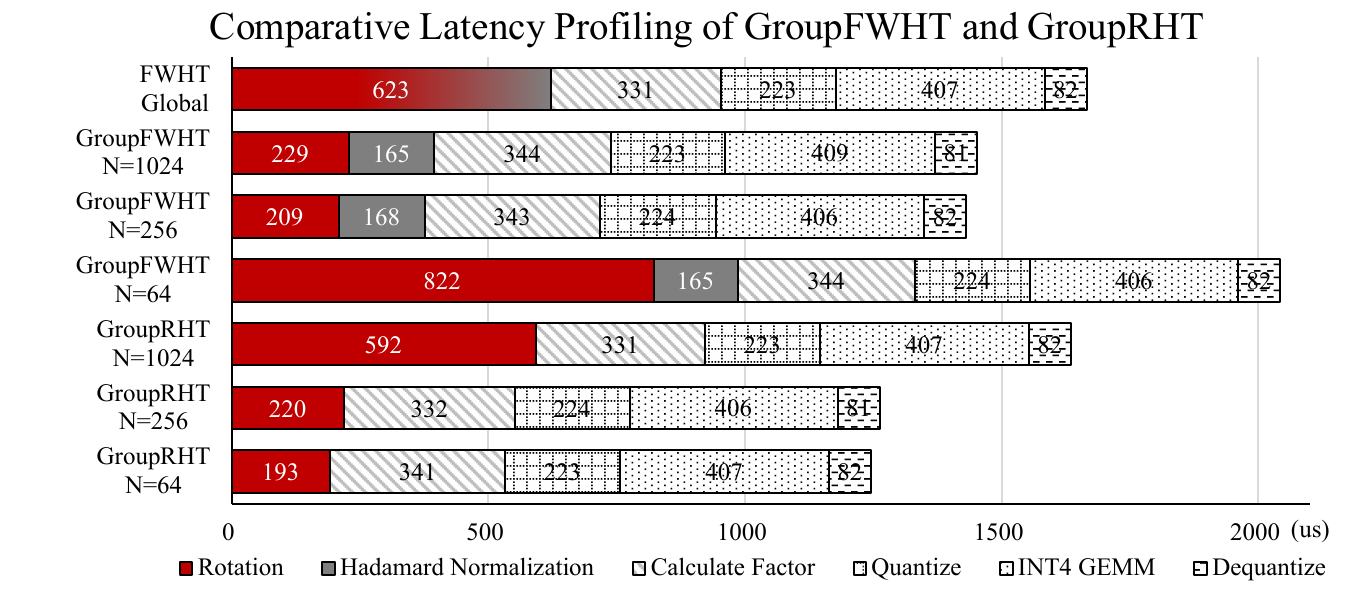}
  \caption{Latency comparison of FWHT and RHT under different group sizes.}
  \lblfig{app-3}
\end{figure}

\section{More Experiment Results}
\lblapp{ablation}

In this section, we present additional experimental results. At the layer level, we show the outlier suppression effect of different Hadamard matrices and group-wise rotations across various linear layers in FLUX.1-dev, demonstrating the effectiveness of ConvRot. At the end-to-end level, we evaluate the impact of different Hadamard matrices, group-wise rotations, and quantization bitwidths on the final generated image quality.

\subsection{Layer-level Outlier Suppression Analysis}
\lblapp{layer}

As shown in \tbl{app-2}, we provide a detailed comparison of three Hadamard matrices and four group sizes across different layers in FLUX.1-dev, using the prompt ``a cute cat''. The results indicate that, in general, larger group sizes $N$ lead to stronger outlier suppression. The Regular matrix exhibits the most stable performance, while the Standard matrix may fail in layers with row-wise outliers, such as \texttt{single\_transformer\_blocks\_\{i\}\_proj\_out}.

\begin{table}[t]
    \centering
    \caption{Outlier amplitude after rotation across transformer layers in FLUX.1-dev. \textbf{Bold} and \underline{underlined} numbers indicate the largest and second-largest reduction (i.e., the strongest suppression effect).}
  
    \lbltbl{app-2}
    \footnotesize
    \setlength{\tabcolsep}{1pt}
   
    \begin{tabular}{l|cccccc}
    \toprule
    \multicolumn{7}{c}{\textit{single\_transformer\_blocks\_0\_attn\_to\_k}} \\
    \midrule
    \multirow{2}{*}{\makecell[l]{Hadamard\\Type}} & \multicolumn{6}{c}{Outlier Amplitude $\downarrow$} \\
    \cmidrule(lr){2-7}
    & $N_0\!=\!16$ & $N_0\!=\!64$ & $N_0\!=\!256$ & $N_0\!=\!1024$ & Global & Original \\
    \midrule
    Random & 18.09{\color{mypink}\scriptsize -57\%} & 12.70{\color{mypink}\scriptsize -69\%} & 8.59{\color{mypink}\scriptsize -79\%} & 5.66{\color{mypink}\scriptsize -86\%} & & 41.62 \\
    Standard & 11.16{\color{mypink}\scriptsize -73\%} & 6.32{\color{mypink}\scriptsize -85\%} & 4.29{\color{mypink}\scriptsize -90\%} & \underline{4.14}{\color{mypink}\scriptsize -90\%} & 4.21{\color{mypink}\scriptsize -90\%} & 41.62 \\
    \rowcolor{gray!12}
    Regular & 10.95{\color{mypink}\scriptsize -74\%} & 6.46{\color{mypink}\scriptsize -84\%} & 4.48{\color{mypink}\scriptsize -89\%} & \textbf{3.54}{\color{mypink}\scriptsize -92\%} & & 41.62 \\
    \midrule
    \multicolumn{7}{c}{\textit{single\_transformer\_blocks\_27\_proj\_mlp}} \\
    \midrule
    \multirow{2}{*}{\makecell[l]{Hadamard\\Type}} & \multicolumn{6}{c}{Outlier Amplitude $\downarrow$} \\
    \cmidrule(lr){2-7}
    & $N_0\!=\!16$ & $N_0\!=\!64$ & $N_0\!=\!256$ & $N_0\!=\!1024$ & Global & Original \\
    \midrule
    Random & 6.84{\color{mypink}\scriptsize -43\%} & 3.71{\color{mypink}\scriptsize -69\%} & 3.00{\color{mypink}\scriptsize -75\%} & 1.97{\color{mypink}\scriptsize -84\%} & & 12.11 \\
    Standard & 3.70{\color{mypink}\scriptsize -69\%} & 2.42{\color{mypink}\scriptsize -80\%} & \underline{1.91}{\color{mypink}\scriptsize -84\%} & 2.12{\color{mypink}\scriptsize -82\%} &\textbf{1.78}{\color{mypink}\scriptsize -85\%} & 12.11 \\
    \rowcolor{gray!12}
    Regular & 3.68{\color{mypink}\scriptsize -70\%} & 2.42{\color{mypink}\scriptsize -80\%} & 2.35{\color{mypink}\scriptsize -81\%} & 2.04{\color{mypink}\scriptsize -83\%} & & 12.11 \\
    \midrule
    \multicolumn{7}{c}{\textit{single\_transformer\_blocks\_24\_proj\_out}} \\
    \midrule
    \multirow{2}{*}{\makecell[l]{Hadamard\\Type}} & \multicolumn{6}{c}{Outlier Amplitude $\downarrow$} \\
    \cmidrule(lr){2-7}
    & $N_0\!=\!16$ & $N_0\!=\!64$ & $N_0\!=\!256$ & $N_0\!=\!1024$ & Global & Original \\
    \midrule
    Random & 3.57{\color{mypink}\scriptsize -53\%} & 2.88{\color{mypink}\scriptsize -62\%} & 2.43{\color{mypink}\scriptsize -68\%} & \textbf{2.14}{\color{mypink}\scriptsize -72\%} & & 7.59 \\
    Standard & 3.27{\color{mypink}\scriptsize -57\%} & 3.88{\color{mypink}\scriptsize -49\%} & 6.45{\color{mypink}\scriptsize -15\%} & 10.79{\color{blue}\scriptsize +42\%} & 30.97{\color{blue}\scriptsize +308\%} & 7.59 \\
    \rowcolor{gray!12}
    Regular & 3.19{\color{mypink}\scriptsize -58\%} & 3.54{\color{mypink}\scriptsize -53\%} & 2.70{\color{mypink}\scriptsize -64\%} & \underline{2.25}{\color{mypink}\scriptsize -70\%} & & 7.59 \\
    \midrule
    \multicolumn{7}{c}{\textit{transformer\_blocks\_12\_ff\_context\_net\_2}} \\ 
    \midrule
    \multirow{2}{*}{\makecell[l]{Hadamard\\Type}} & \multicolumn{6}{c}{Outlier Amplitude $\downarrow$} \\
    \cmidrule(lr){2-7}
    & $N_0\!=\!16$ & $N_0\!=\!64$ & $N_0\!=\!256$ & $N_0\!=\!1024$ & Global & Original \\
    \midrule
    Random & 16.94{\color{mypink}\scriptsize -27\%} & 14.25{\color{mypink}\scriptsize -38\%} & 10.38{\color{mypink}\scriptsize -55\%} & \underline{9.52}{\color{mypink}\scriptsize -59\%} & & 23.11 \\ 
    Standard & 13.23{\color{mypink}\scriptsize -43\%} & 10.60{\color{mypink}\scriptsize -54\%} & 12.70{\color{mypink}\scriptsize -45\%} & 20.39{\color{mypink}\scriptsize -12\%} & 43.81{\color{blue}\scriptsize +90\%} & 23.11 \\ 
    \rowcolor{gray!12}
    Regular & 13.23{\color{mypink}\scriptsize -43\%} & 10.65{\color{mypink}\scriptsize -54\%} & \textbf{8.99}{\color{mypink}\scriptsize -61\%} & 9.74{\color{mypink}\scriptsize -58\%} & & 23.11 \\ 
    \midrule
    \multicolumn{7}{c}{\textit{transformer\_blocks\_3\_attn\_to\_out\_0}} \\
    \midrule
    \multirow{2}{*}{\makecell[l]{Hadamard\\Type}} & \multicolumn{6}{c}{Outlier Amplitude $\downarrow$} \\
    \cmidrule(lr){2-7}
    & $N_0\!=\!16$ & $N_0\!=\!64$ & $N_0\!=\!256$ & $N_0\!=\!1024$ & Global & Original \\
    \midrule
    Random & 6.56{\color{blue}\scriptsize +2\%} & 5.65{\color{mypink}\scriptsize -12\%} & 4.96{\color{mypink}\scriptsize -23\%} & 2.79{\color{mypink}\scriptsize -57\%} & & 6.45 \\
    Standard & 6.87{\color{blue}\scriptsize +7\%} & 6.05{\color{mypink}\scriptsize -6\%} & 4.06{\color{mypink}\scriptsize -37\%} & 2.75{\color{mypink}\scriptsize -57\%} & \textbf{2.41}{\color{mypink}\scriptsize -63\%} & 6.45 \\
    \rowcolor{gray!12}
    Regular & 5.57{\color{mypink}\scriptsize -14\%} & 6.76{\color{blue}\scriptsize +5\%} & 4.38{\color{mypink}\scriptsize -32\%} & \underline{2.51}{\color{mypink}\scriptsize -61\%} & & 6.45 \\
    \midrule
    \multicolumn{7}{c}{\textit{transformer\_blocks\_2\_ff\_net\_2}} \\ 
    \midrule
    \multirow{2}{*}{\makecell[l]{Hadamard\\Type}} & \multicolumn{6}{c}{Outlier Amplitude $\downarrow$} \\
    \cmidrule(lr){2-7}
    & $N_0\!=\!16$ & $N_0\!=\!64$ & $N_0\!=\!256$ & $N_0\!=\!1024$ & Global & Original \\
    \midrule
    Random & 1.96{\color{mypink}\scriptsize -37\%} & 1.27{\color{mypink}\scriptsize -59\%} & 1.02{\color{mypink}\scriptsize -67\%} & \underline{0.94}{\color{mypink}\scriptsize -70\%} & & 3.10 \\ 
    Standard & 1.28{\color{mypink}\scriptsize -59\%} & 1.22{\color{mypink}\scriptsize -61\%} & 1.36{\color{mypink}\scriptsize -56\%} & 2.15{\color{mypink}\scriptsize -30\%} & 4.94{\color{blue}\scriptsize +60\%} & 3.10 \\
    \rowcolor{gray!12}
    Regular & 1.17{\color{mypink}\scriptsize -62\%} & 0.99{\color{mypink}\scriptsize -68\%} & \textbf{0.94}{\color{mypink}\scriptsize -70\%} & 1.00{\color{mypink}\scriptsize -68\%} & & 3.10 \\ 
    \bottomrule
    \end{tabular}
  \end{table}

\subsection{End-to-End Image Generation Ablation}

As shown in \tbl{app-3}, we compare the end-to-end generation quality obtained using the Regular Hadamard matrix versus the Standard Hadamard matrix. To evaluate the effect of block size, we modify QuaRot by replacing its standard full-rank FWHT with a block-wise FWHT, denoted as QuaRot$^{\dagger}$. The results show that under the W4A4 setting, QuaRot with a global Standard Hadamard transform completely fails. Under local block-wise settings, using the Regular Hadamard matrix yields noticeably better performance than using the Standard version, whereas the gap becomes less pronounced in the W8A8 setting due to the reduced difficulty of 8-bit quantization. These findings indicate that simply adopting a block-wise strategy already mitigates much of the outlier concentration effect caused by the Standard Hadamard matrix, while switching to the Regular matrix eliminates this issue entirely. It is worth noting that we do not evaluate a global RHT, as there is currently no known construction method for Regular matrices of arbitrary sizes.  Our proposed construction Theorem~\ref{thm:kronecker} only applies to sizes that are powers of four, and therefore can be used only for group-wise local transforms rather than global ones.

\begin{table*}[]
    \footnotesize
      \setlength{\tabcolsep}{7pt}
      \centering
      \caption{
      Ablation on the choice of rotation type and size. We compare Standard and Regular Hadamard transforms under global and block-wise settings across different quantization precisions. Mixed refers to the use of a mixed precision strategy, where approximately 20\% of the linear layers are quantized to INT 8.
  }
  \lbltbl{app-3}
    \begin{tabular}{ccccccccccccc}
      \toprule
     &  &  & \multicolumn{2}{c}{Rotation} & \multicolumn{4}{c}{MJHQ} & \multicolumn{4}{c}{sDCI} \\
     \cmidrule(lr){4-5} \cmidrule(lr){6-9} \cmidrule(lr){10-13}
    \multirow{-2}{*}{Model} & \multirow{-2}{*}{Precision} & \multirow{-2}{*}{Mixed} & Type & Size & FID$\downarrow$ & IR$\uparrow$ & LPIPS$\downarrow$ & PSNR$\uparrow$ & FID$\downarrow$ & IR$\uparrow$ & LPIPS$\downarrow$ & PSNR$\uparrow$ \\
    \midrule
     & BF16 &  & -- & -- & 10.07 & 0.993 & -- & -- & 13.84 & 1.055 & -- & -- \\
     \cmidrule(lr){2-13}
     & INT W8A8 &  & Standard & 256 & 9.93 & 0.981 & 0.140 & 22.64 & 13.45 & 1.018 & 0.157 & 22.60 \\
     & INT W8A8 &  & Standard & 1024 & 9.91 & 0.985 & 0.139 & 22.61 & 13.50 & 1.001 & 0.166 & 22.69 \\
     & \cellcolor{mygray}INT W8A8 & \cellcolor{mygray} & \cellcolor{mygray}Regular & \cellcolor{mygray}256 & \cellcolor{mygray}9.82 & \cellcolor{mygray}0.981 & \cellcolor{mygray}0.131 & \cellcolor{mygray}22.62 & \cellcolor{mygray}13.47 & \cellcolor{mygray}1.050 & \cellcolor{mygray}0.157 & \cellcolor{mygray}22.59 \\
     & \cellcolor{mygray}INT W8A8 & \cellcolor{mygray} & \cellcolor{mygray}Regular & \cellcolor{mygray}1024 & \cellcolor{mygray}9.77 & \cellcolor{mygray}0.977 & \cellcolor{mygray}0.129 & \cellcolor{mygray}22.66 & \cellcolor{mygray}13.46 & \cellcolor{mygray}1.044 & \cellcolor{mygray}0.148 & \cellcolor{mygray}22.62 \\
     \cmidrule(lr){2-13}
     & {\color[HTML]{BFBFBF} INT W4A4} & {\color[HTML]{BFBFBF} \ding{51}} & {\color[HTML]{BFBFBF} Standard} & {\color[HTML]{BFBFBF} Full} & {\color[HTML]{BFBFBF} 98.04} & {\color[HTML]{BFBFBF} -2.240} & {\color[HTML]{BFBFBF} 0.766} & {\color[HTML]{BFBFBF} 7.70} & {\color[HTML]{BFBFBF} 110.84} & {\color[HTML]{BFBFBF} -2.277} & {\color[HTML]{BFBFBF} 0.767} & {\color[HTML]{BFBFBF} 7.41} \\
     & INT W4A4 &  & Standard & 256 & 12.55 & 0.835 & 0.234 & 19.23 & 14.09 & 0.994 & 0.247 & 17.89 \\
     & INT W4A4 &  & Standard & 1024 & 13.86 & 0.808 & 0.286 & 18.85 & 16.99 & 0.913 & 0.345 & 16.83 \\
     & INT W4A4 & \ding{51} & Standard & 256 & 12.45 & 0.861 & 0.193 & 20.17 & 15.86 & 0.878 & 0.241 & 19.24 \\
     & INT W4A4 & \ding{51} & Standard & 1024 & 13.29 & 0.850 & 0.290 & 18.45 & 16.53 & 0.900 & 0.360 & 16.22 \\
     & \cellcolor{mygray}INT W4A4 & \cellcolor{mygray} & \cellcolor{mygray}Regular & \cellcolor{mygray}256 & \cellcolor{mygray}12.32 & \cellcolor{mygray}0.841 & \cellcolor{mygray}0.220 & \cellcolor{mygray}19.43 & \cellcolor{mygray}16.02 & \cellcolor{mygray}0.870 & \cellcolor{mygray}0.249 & \cellcolor{mygray}17.66 \\
 & \cellcolor{mygray}INT W4A4 & \cellcolor{mygray} & \cellcolor{mygray}Regular & \cellcolor{mygray}1024 & \cellcolor{mygray}12.30 & \cellcolor{mygray}0.855 & \cellcolor{mygray}0.215 & \cellcolor{mygray}19.52 & \cellcolor{mygray}15.87 & \cellcolor{mygray}0.880 & \cellcolor{mygray}0.245 & \cellcolor{mygray}17.72 \\
 & \cellcolor{mygray}INT W4A4 & \cellcolor{mygray}\ding{51} & \cellcolor{mygray}Regular & \cellcolor{mygray}256 & \cellcolor{mygray}10.03 & \cellcolor{mygray}0.973 & \cellcolor{mygray}0.186 & \cellcolor{mygray}20.73 & \cellcolor{mygray}14.00 & \cellcolor{mygray}1.020 & \cellcolor{mygray}0.214 & \cellcolor{mygray}19.41 \\
\multirow{-16}{*}{\begin{tabular}[c]{@{}c@{}}FLUX.1\\      -dev\\      (50 Steps)\end{tabular}} & \cellcolor{mygray}INT W4A4 & \cellcolor{mygray}\ding{51} & \cellcolor{mygray}Regular & \cellcolor{mygray}1024 & \cellcolor{mygray}\textbf{10.02} & \cellcolor{mygray}0.983 & \cellcolor{mygray}0.184 & \cellcolor{mygray}20.71 & \cellcolor{mygray}14.02 & \cellcolor{mygray}1.020 & \cellcolor{mygray}0.224 & \cellcolor{mygray}19.42 \\
    \midrule
     & BF16 &  & -- & -- & 11.59 & 0.915 & -- & -- & 11.01 & 0.974 & -- & -- \\
     \cmidrule(lr){2-13}
     & INT W8A8 &  & Standard & 256 & 10.90 & 0.962 & 0.151 & 21.36 & 11.28 & 1.006 & 0.152 & 19.93 \\
     & INT W8A8 &  & Standard & 1024 & 10.88 & 0.971 & 0.142 & 21.68 & 11.25 & 1.002 & 0.155 & 20.15 \\
     & \cellcolor{mygray}INT W8A8 & \cellcolor{mygray} & \cellcolor{mygray}Regular & \cellcolor{mygray}256 & \cellcolor{mygray}10.86 & \cellcolor{mygray}0.958 & \cellcolor{mygray}0.151 & \cellcolor{mygray}21.36 & \cellcolor{mygray}11.09 & \cellcolor{mygray}0.995 & \cellcolor{mygray}0.157 & \cellcolor{mygray}19.91 \\
     & \cellcolor{mygray}INT W8A8 & \cellcolor{mygray} & \cellcolor{mygray}Regular & \cellcolor{mygray}1024 & \cellcolor{mygray}10.84 & \cellcolor{mygray}0.965 & \cellcolor{mygray}0.151 & \cellcolor{mygray}21.39 & \cellcolor{mygray}11.05 & \cellcolor{mygray}1.007 & \cellcolor{mygray}0.136 & \cellcolor{mygray}20.11 \\    
     \cmidrule(lr){2-13}
     & {\color[HTML]{BFBFBF} INT W4A4} & {\color[HTML]{BFBFBF} \ding{51}} & {\color[HTML]{BFBFBF} Standard} & {\color[HTML]{BFBFBF} Full} & {\color[HTML]{BFBFBF} 101.09} & {\color[HTML]{BFBFBF} -2.252} & {\color[HTML]{BFBFBF} 0.811} & {\color[HTML]{BFBFBF} 6.54} & {\color[HTML]{BFBFBF} 114.01} & {\color[HTML]{BFBFBF} -2.279} & {\color[HTML]{BFBFBF} 0.601} & {\color[HTML]{BFBFBF} 6.75} \\
     & INT W4A4 &  & Standard & 256 & 14.47 & 0.803 & 0.242 & 17.59 & 13.54 & 0.894 & 0.313 & 16.88 \\
     & INT W4A4 &  & Standard & 1024 & 16.19 & 0.761 & 0.259 & 16.44 & 13.86 & 0.910 & 0.353 & 16.47 \\
     & INT W4A4 & \ding{51} & Standard & 256 & 12.13 & 0.883 & 0.204 & 18.05 & 12.41 & 0.965 & 0.281 & 18.66 \\
     & INT W4A4 & \ding{51} & Standard & 1024 & 14.07 & 0.890 & 0.290 & 16.74 & 13.22 & 0.970 & 0.270 & 17.80 \\
     & \cellcolor{mygray}INT W4A4 & \cellcolor{mygray} & \cellcolor{mygray}Regular & \cellcolor{mygray}256 & \cellcolor{mygray}13.38 & \cellcolor{mygray}0.814 & \cellcolor{mygray}0.231 & \cellcolor{mygray}17.50 & \cellcolor{mygray}12.37 & \cellcolor{mygray}0.844 & \cellcolor{mygray}0.301 & \cellcolor{mygray}16.98 \\
     & \cellcolor{mygray}INT W4A4 & \cellcolor{mygray} & \cellcolor{mygray}Regular & \cellcolor{mygray}1024 & \cellcolor{mygray}12.88 & \cellcolor{mygray}0.868 & \cellcolor{mygray}0.224 & \cellcolor{mygray}17.73 & \cellcolor{mygray}13.59 & \cellcolor{mygray}0.949 & \cellcolor{mygray}0.314 & \cellcolor{mygray}16.92 \\
     & \cellcolor{mygray}INT W4A4 & \cellcolor{mygray}\ding{51} & \cellcolor{mygray}Regular & \cellcolor{mygray}256 & \cellcolor{mygray}11.49 & \cellcolor{mygray}0.926 & \cellcolor{mygray}0.201 & \cellcolor{mygray}18.11 & \cellcolor{mygray}11.13 & \cellcolor{mygray}0.975 & \cellcolor{mygray}0.228 & \cellcolor{mygray}18.71 \\
    \multirow{-16}{*}{\begin{tabular}[c]{@{}c@{}}FLUX.1\\      -schnell\\      (50 Steps)\end{tabular}} & \cellcolor{mygray}INT W4A4 & \cellcolor{mygray}\ding{51} & \cellcolor{mygray}Regular & \cellcolor{mygray}1024 & \cellcolor{mygray}11.27 & \cellcolor{mygray}0.909 & \cellcolor{mygray}0.203 & \cellcolor{mygray}18.06 & \cellcolor{mygray}11.31 & \cellcolor{mygray}0.995 & \cellcolor{mygray}0.245 & \cellcolor{mygray}18.68 \\    
    \bottomrule
    \end{tabular}
    \end{table*}

As shown in \fig{app-4}, we evaluate the impact of quantization bitwidth on end-to-end image generation quality. By default, ConvRot quantizes all linear layers in the (single) transformer blocks, except for \texttt{transformer\_blocks\_18\_ff\_context\_net\_2}, \texttt{transformer\_blocks\_18\_ff\_net\_2} and \texttt{single\_transformer\_blocks.37.proj\_out}\footnote{These layers are the final layers of the \texttt{transformer\_blocks} and \texttt{single\_transformer\_blocks}, which significantly affect image quality. Experiments show that keeping them at higher precision has minimal impact on model speed after quantization.}.  

Under the W8A8 setting, ConvRot achieves good image quality. Under W4A4, ConvRot exhibits the best inference speed with a moderate and acceptable degradation in image quality. By applying a mixed-precision strategy, using 8-bit mixed precision for 20\% of sensitive layer, the degradation can be effectively mitigated, improving overall image quality. Details of the mixed-precision strategy can be found in \app{mix}.

\fig{app-5} further illustrates the degradation pattern observed under W4A4. Although high-frequency details such as clothing wrinkles and hair are preserved, low-frequency regions exhibit noticeable mosaicking. We rule out the influence of outlier suppression, since restoring selected sensitive layers to int8 recovers low-frequency details, even though these layers do not exhibit significant outlier activations (e.g., \texttt{transformer\_blocks\_\{i\}\_attn\_to\_out\_0}).  

We hypothesize that this is due to the limited representational range of 4-bit quantization, which makes it difficult for the model to capture subtle transitions. Such degradation commonly occurs in smooth regions, such as skies or walls.

\begin{figure}[h]
    \centering
    \includegraphics[width=1\linewidth]{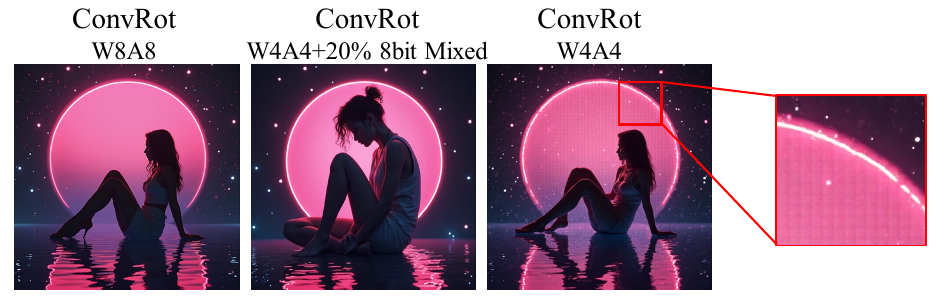}
    \caption{Visualization of image degradation under low-bit quantization.}
    \lblfig{app-5}
\end{figure}

\begin{figure*}[b]
    \centering
    \includegraphics[width=0.7\linewidth]{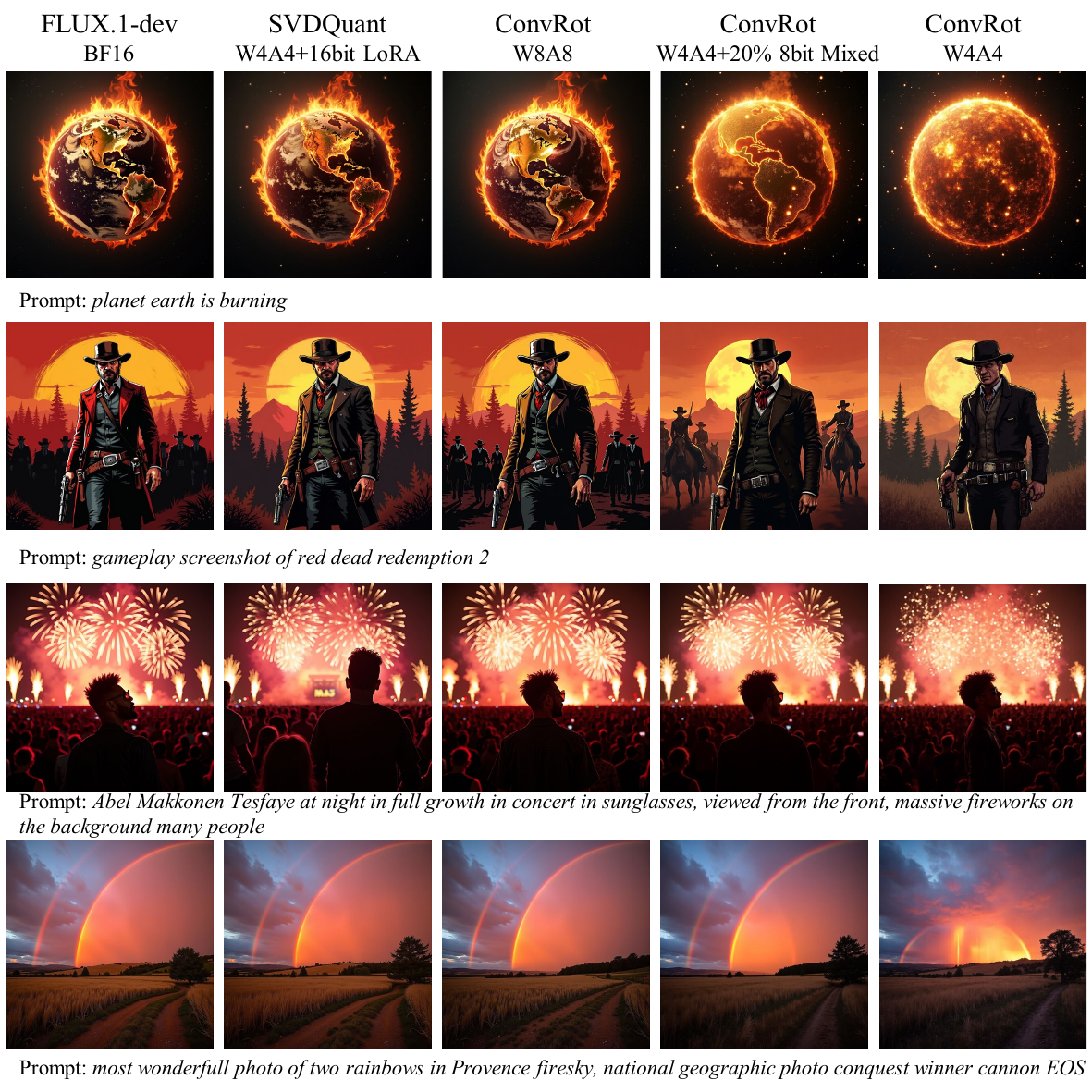}
    \caption{Ablation study on the impact of quantization bitwidth on end-to-end image generation quality.}
    \lblfig{app-4}
\end{figure*}

\section{Mixed-Precision Strategy}
\lblapp{mix}

\begin{table}[b]
\centering
\small
\caption{Layer-wise mixed-precision quantization list.}
\lbltbl{app-4}
\begin{tabular}{ll}
\toprule
\textbf{Layer} & \textbf{Precision} \\
\midrule
transformer\_blocks\_\{i\}\_attn\_to\_out\_0 & W8A8 \\
single\_transformer\_blocks\_\{i\}\_attn\_to\_v & W8A8 \\
single\_transformer\_blocks\_37\_proj\_out & W8A8 \\
transformer\_blocks\_18\_ff\_context\_net\_2 & W8A8 \\ 
transformer\_blocks\_18\_ff\_net\_2 & W8A8 \\ \hline
others & W4A4 \\
\bottomrule
\end{tabular}
\end{table}

As shown in \fig{app-5}, we observe noticeable image degradation under the W4A4 setting. We hypothesize that the loss of low-frequency details arises from the limited representational capacity of W4A4, which struggles to capture subtle transitions. To compensate for this deficiency, we introduce additional expressiveness into the quantized model.

Inspired by the 16-bit LoRA used in SVDQuant, we adopt a mixed-precision strategy: based on empirical observations, we manually quantize layers one by one and find that a subset of layers has a disproportionately large impact on image quality. These layers are therefore assigned higher precision (W8A8 instead of W4A4) to improve overall fidelity rather than being kept in FP16/BF16. Notably, during this process we did not observe strong outlier activations in these sensitive layers, suggesting that their importance is tied more to their functional role in the network than to activation statistics.

In practice, we select approximately 20\% of the linear layers for W8A8 quantization; the exact layer list is provided in \tbl{app-4}.

\section{Comparison with Existing Quantization Methods}
\lblapp{compare}

\myparagraph{Comparison with QuaRot.}
QuaRot~\citep{ashkboos2024quarot} is a calibration-free PTQ method designed for LLMs, reducing rotation overhead by fusing rotations into weights and accelerating them via FWHT. However, its rotation-reduction strategy does not apply to DiT, and the FWHT induces row-wise outlier aggregation. Building upon QuaRot, we retain its quantization, INT4 GEMM, and dequantization kernels while introducing our Group RHT within a unified ConvLinear4bit layer. Group-wise rotation lowers computational cost and mitigates DiT's outlier amplification, and the use of regular Hadamard matrices eliminates the amplification issue entirely.

\myparagraph{Comparison with NF4.}
NF4~\citep{dettmers2023qlora} is the 4-bit NormalFloat data type introduced in QLoRA, designed to be information-theoretically optimal for normally distributed weights and effective for low-precision finetuning. However, NF4 requires dequantization into higher-precision formats during computation and therefore provides limited inference-time acceleration when applied to diffusion transformers. In contrast, our ConvRot method enables direct INT4 computation with integrated rotation and quantization, achieving significant speedups during inference.

\myparagraph{Comparison with DuQuant.}
DuQuant~\citep{lin2024duquant} also explores block-wise rotation for LLMs, targeting massive activations by permuting affected channels and applying standard Hadamard transforms inside each block. Its complex matrix construction (SmoothQuant + permutation + two rotations), however, offers limited speed benefits. In contrast, our Group RHT is tailored to DiT's row-wise outlier structure, avoids heavy matrix design, and enables a substantially faster rotation procedure.

\myparagraph{Comparison with SVDQuant.}
SVDQuant~\citep{li2024svdquant} is the first to achieve W4A4 quantization on DiT, delivering strong image quality and impressive speed through comprehensive engineering optimizations. We highly appreciate the authors' engagement with the open-source community, its out-of-the-box usability makes reproduction and comparison straightforward. However, its reliance on the specialized nunchaku inference engine restricts broader applicability. In contrast, our method is lightweight and easy to integrate, requires no additional runtime dependencies, and supports plug-and-play DiT quantization with competitive performance.

\myparagraph{Comparison with HadaNorm.}
HadaNorm~\citep{federici2025hadanorm} attributes the failure of standard Hadamard transforms to mean and scale differences across channels, where the all-ones row prevents non-zero-mean channels from avoiding long-tailed activations; they address this via explicit mean subtraction. Our approach resolves the issue directly through rotation-matrix design, eliminating extra steps. As HadaNorm is not yet open-sourced and reports limited experiments, we omit a direct comparison. Incidentally, we were unaware of this work when developing our method; after arriving at our approach independently, we were pleased to find that other researchers had identified a similar problem and proposed an alternative, effective solution.

\end{document}